\documentclass[letterpaper, 10 pt, journal]{ieeeconf}  

\IEEEoverridecommandlockouts                              

\usepackage{graphicx} 
\usepackage{amsmath} 
\usepackage{amssymb}  
\usepackage{hhline}
\usepackage[colorlinks=true,linkcolor=blue]{hyperref} 
\usepackage{tikz}
\usepackage{hypcap} 
\usepackage{xcolor} 
\usepackage{soul}   
\usepackage{transparent}

\newcommand{\revised}[1]{#1}
\newcommand{\revisionTwo}[1]{#1}

\newcommand*\circled[1]{\tikz[baseline=(char.base)]{
            \node[shape=circle,draw,inner sep=1pt] (char) {#1};}}

\title{
They See Me Rolling: High-Speed Event Vision-Based Tactile Roller Sensor for Large Surface Inspection
}

\author{Akram Khairi$^{1}$, Hussain Sajwani$^{1}$, Abdallah Mohammad Alkilany$^{1}$, Laith AbuAssi$^{1}$, Mohamad Halwani$^{1}$, Islam Mohamed Zaid$^{1}$, Ahmed Awadalla$^{1}$, Dewald Swart$^{3}$, Abdulla Ayyad$^{1}$, Yahya Zweiri$^{1,2}$ 
\thanks{This research was funded by Sandooq Al Watan under Project ID: KU-EXT-SWARD-2022-8434000486. The authors acknowledge the support of the Advanced Research and Innovation Center (KU-ARIC), a joint research center established by Khalifa University of Science and Technology and Aerospace Holding Company LLC, a wholly-owned subsidiary of Mubadala Investment Company PJSC. Akram Khairi is the corresponding author (\emph{email: akram.hassan@ku.ac.ae}).}
\thanks{$^{1}$A. Khairi, H. Sajwani, A. M. Alkilany, L. AbuAssi, M. Halwani, I. M. Zaid, A. Awadalla, A. Ayyad and Y. Zweiri are with the Advanced Research and Innovation Center (ARIC), Khalifa University, Abu Dhabi, United Arab Emirates. 
$^{2}$Y. Zweiri is also with the Department of Aerospace Engineering, Khalifa University, Abu Dhabi, United Arab Emirates. 
$^{3}$D. Swart is with Research and Development, Strata Manufacturing PJSC, Al Ain, United Arab Emirates.}%
}

\begin{document}

\setlength{\fboxsep}{2pt}%
\setlength{\fboxrule}{0.5pt}%

\maketitle
\thispagestyle{empty}
\pagestyle{empty}

\begin{abstract}
Inspecting large-scale industrial surfaces like aircraft fuselages for quality control requires precise, high-resolution 3D geometry. Vision-based tactile sensors (VBTSs) offer high local resolution but require slow 'press-and-lift' measurements for large areas. Sliding or roller/belt VBTS designs provide continuous measurement but face significant challenges: sliding suffers from friction/wear, while both are speed-limited by camera frame rates and motion blur. Thus, a rapid, continuous, high-resolution method is needed. We introduce a novel neuromorphic tactile roller sensor. It uses a modified event-based multi-view stereo algorithm for 3D reconstruction, leveraging high temporal resolution and motion blur robustness. \revised{This reconstruction is most effective for surfaces with distinct edges or sharp features, which are often the most critical for defect detection in industrial inspection tasks.} We demonstrate 0.5 m/s scanning speeds with MAE below 100 µm (11x faster than prior methods). A multi-reference Bayesian fusion strategy reduces MAE by 25.2\% (vs. EMVS) and mitigates curvature errors. We also validate high-speed Braille reading 2.6x faster than previous approaches.
\end{abstract}

    \section{Introduction}
Surface metrology and surface inspection are crucial elements in quality assurance across diverse industries, particularly aerospace and automotive manufacturing. Precise inspection is required to identify characteristics like paint quality, coating integrity, and subtle defects such as cracks, nicks, and dents \cite{futterlieb2017air, yusra, Yuan2017}. Often, achieving a resolution of 0.1 mm or lower is necessary to accurately classify these features and ensure component integrity and safety \cite{Agarwal2023}. Traditional contact-based methods, including high-precision profilometers \cite{sekatskii2019analysis, Villarrubia1997} or microscopic techniques \cite{Russell2001, Yang2018, Jonkman2020}, offer high resolution locally but become exceedingly time-consuming when applied to large surface areas due to their sequential, point-by-point or small-patch measurement nature. Non-contact optical methods, such as cameras, laser scanners, or structured light systems \cite{yusra, Salah2023, deng2024review, laser, zhang2018high,ramadan_vision-guided_2024}, can significantly accelerate inspection by capturing data over wider areas. However, these methods often lack robustness; their performance can be compromised by variations in ambient lighting, motion blur when attempting high-speed scanning, or challenging surface optical properties like high reflectivity or transparency \cite{Hausler2011}. This highlights a critical need for sensing modalities that combine high resolution and robustness with the capability for rapid inspection of large, potentially complex surfaces. \revised{Capturing the full 3D geometry is particularly important, as it enables the quantification of volumetric defects (e.g., dent depth and volume) and allows for direct comparison against a target CAD model, a critical step in modern quality assurance.}

Vision-Based Tactile Sensors (VBTSs), particularly those derived from the GelSight concept \cite{Yuan2017, Johnson2011, li2020review}, have emerged as a powerful contact-based alternative offering a unique blend of high-resolution geometric and textural sensing. These sensors typically employ a camera to view the deformation of a soft, reflective elastomer membrane pressed against a target surface. This principle allows for detailed surface capture, often with inherent robustness to external lighting conditions due to internal illumination \cite{Yuan2017, Dong2017}. Numerous variations of GelSight have been developed, enhancing capabilities for robotic manipulation, complex shape capture, and integration into compact form factors \cite{Donlon2018, Liu2022, Zhao2023, Jing2025}.

Despite their advantages for detailed local analysis, the standard planar configuration of most VBTSs necessitates a discrete "press-and-lift" interaction mode. This process, where the sensor is repeatedly pressed, lifted, and moved, becomes prohibitively time-consuming when inspecting large surface areas, requiring numerous measurements to be stitched together \cite{Agarwal2023}, thereby limiting their use for efficient large-scale tactile measurement.

\begin{figure*}[t] \centering \includegraphics[width = \linewidth]{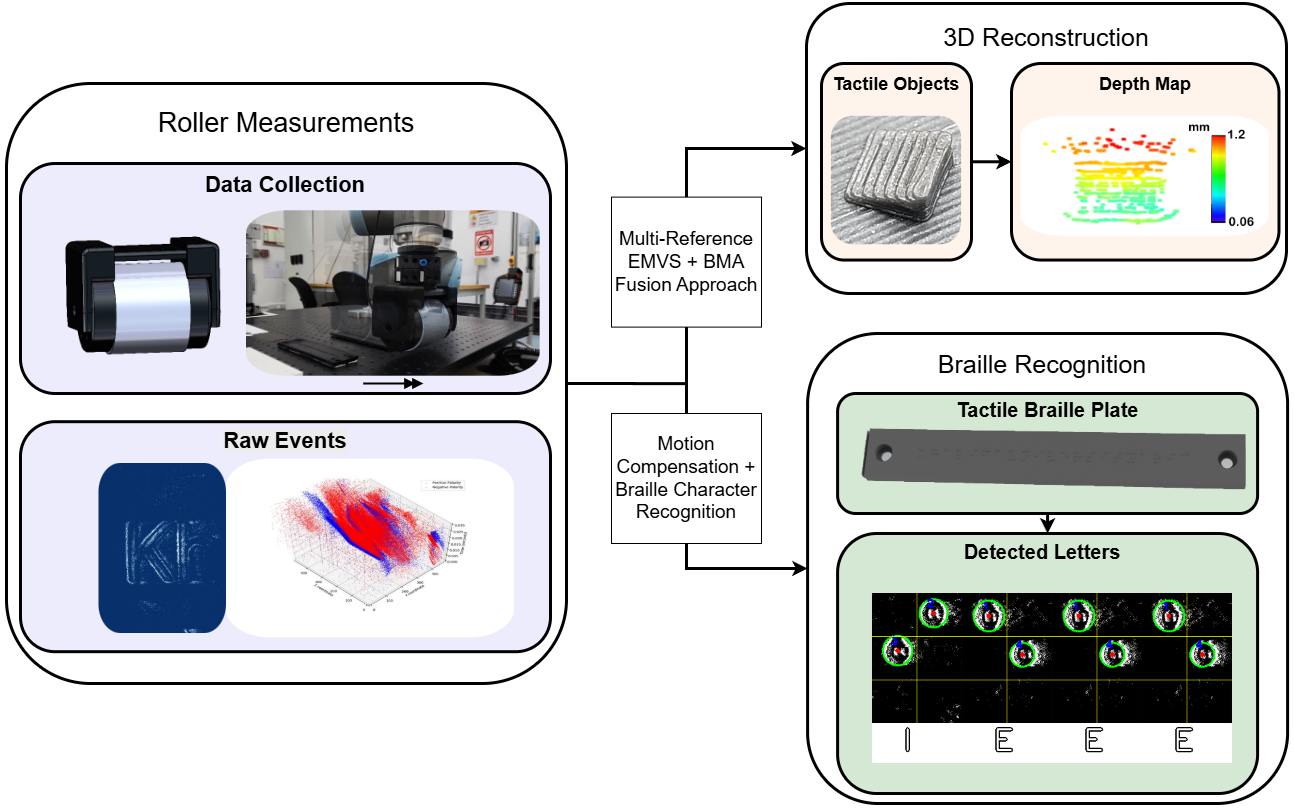} \caption{The proposed neuromorphic roller tactile sensor enables high-speed surface analysis. The system (mounted on UR10 robotic arm) captures asynchronous events during continuous rolling (left) and utilizes event-based algorithms for both 3D surface reconstruction (top right) and fine feature recognition, demonstrated via Braille reading (bottom right), surpassing the speed limitations of conventional tactile methods for large surface inspection.} \label{fig:graphical_abstract} \end{figure*}

\revised{Neuromorphic vision sensors, or event cameras, provide a promising technological solution to overcome the speed and motion blur limitations of conventional VBTSs \cite{Gallego2022, Huang2022}.} Unlike frame-based cameras, event cameras feature pixels that respond independently and asynchronously to logarithmic brightness changes, generating data with microsecond-level temporal resolution, high dynamic range, and inherent resistance to motion blur \cite{Brandli2014, Posch2011, Lichtsteiner2008}. These characteristics are ideally suited for capturing the rapid, dynamic deformations occurring at the contact interface of a high-speed tactile sensor. Crucially, integrating event cameras into tactile sensing holds considerable potential for achieving high-resolution continuous tactile sensing at higher speeds with significantly lower computational power, particularly for surfaces with sparse details, compared to conventional VBTSs. While Neuromorphic VBTSs (NVBTS) have been explored for applications like robotic grasping and slip detection (e.g., Evetac \cite{Funk2024}) or contact analysis \cite{Ward-Cherrier2020, Sajwani2023, Faris2023, neutac_salah_2024}, their potential for enabling high-speed, continuous 3D large surface reconstruction has not been realized.

\revised{To enable continuous tactile measurement, various designs moving beyond the static planar configuration have emerged.} Attempts at direct sliding of planar sensors face significant friction, wear, and motion blur challenges \cite{potdar_high-speed_2024}. Roller-based \cite{Cao2023, Shimonomura2021} and belt-based \cite{mirzaee_gelbelt_2025} VBTS designs replace the static contact with a dynamic rolling interface, allowing continuous data acquisition. TouchRoller, for example, demonstrated 2D texture mapping over moderate areas \cite{Cao2023}, while GelBelt provides a larger contact patch via its belt mechanism \cite{mirzaee_gelbelt_2025}. Despite these mechanical advancements for measurement continuity, these designs face limitations. Reconstructing accurate 3D geometry from the depth-variant information gathered by cylindrical rollers is complex, primarily because the contact depth is not uniform across the curved sensing area \cite{Li2023}. More critically, the operational speed of systems relying on conventional cameras is constrained. Frame rate limitations and motion blur become significant issues as scanning speed increases, particularly for VBTSs that require close-up imaging with a shallow depth of field to resolve fine surface details. Consequently, current roller/belt tactile sensors are restricted to relatively low velocities (e.g., typically well below 0.1 m/s for 3D reconstruction \cite{Cao2023, mirzaee_gelbelt_2025}), insufficient for truly high-throughput industrial inspection of large parts. A significant gap exists for a tactile sensing system capable of combining the continuous scanning geometry of a roller with the high temporal fidelity and motion robustness of event-based vision for large-scale surface measurement.

This paper introduces, to the best of our knowledge, the first high-speed neuromorphic vision-based roller tactile sensor specifically designed for large surface measurements (illustrated in Figure \ref{fig:graphical_abstract}). \revised{The sensor is effective for 3D reconstruction of surfaces with distinct edges or sharp features, which are often the most critical for tactile feature recognition and defect detection.}
Our primary contributions are: \begin{enumerate}
    \item We present the first work demonstrating the use of a neuromorphic vision-based tactile sensor (NVBTS) for high-resolution 3D reconstruction and surface inspection.
    \revised{
    \item We introduce a novel rolling neuromorphic vision-based tactile sensor for 3D reconstruction and fine feature recognition of surfaces at speeds up to \(0.5\) m/s, 11 times faster than previous continuous tactile sensing approaches. Furthermore, we rigorously evaluate its performance against the state-of-the-art in large surface inspection in terms of both speed and accuracy.
    \item We adapt the event-based multi-view stereo (EMVS) method for curved tactile sensing by integrating a Bayesian Model Averaging (BMA) multi-reference depth fusion strategy, improving depth consistency and reducing curvature-induced errors. Moreover, a corresponding calibration routine is developed to optimize its parameters.
\end{enumerate}}

\section{Related Works}

\subsection{3D Surface Reconstruction}

Reconstructing a surface's three-dimensional geometry is crucial in many applications, from automated manufacturing to quality control. Mechanical profilometers have long been used for high-precision surface measurements, meticulously tracing the surface point-by-point. However, this inherent sequential nature makes them slow, especially when dealing with complex geometries or large areas \cite{Villarrubia1997}. Non-contact optical techniques, like laser scanning and structured light, offer a significant speed advantage by rapidly capturing surface data over a wider area \cite{Salvi2004, Sansoni2009}. These methods boast resolutions around 0.1 mm \cite{Johnson2011}, but their reliance on light makes them susceptible to issues with highly reflective or transparent materials, as well as variations in ambient lighting conditions \cite{Hausler2011}. Furthermore, certain laser scanning systems, particularly higher-power ones used for speed or range, can introduce eye-safety concerns requiring adherence to safety standards and protocols \cite{IEC60825}.

For applications demanding the utmost precision, techniques like white light interferometry, confocal microscopy, and atomic force microscopy (AFM) are utilized \cite{Russell2001, Yang2018, Jonkman2020}. These methods can resolve surface details down to the nanometer scale, revealing intricate textures and features. However, they often involve complex setups, lengthy measurement times, and significant expense, limiting their practical use for large-scale inspection tasks.

\subsection{\revised{Vision-Based and Neuromorphic Tactile Sensing}}

Vision-Based Tactile Sensors (VBTSs) have emerged as compelling alternatives for surface sensing, offering robustness, versatility, and ease of use. The GelSight sensor pioneered this field, making use of a soft, deformable surface with camera-based observation to extract detailed surface information utilizing photometric stereo \cite{Johnson2011}. By analyzing the deformation of the elastomer, typically coated with a reflective layer, VBTSs can capture high-resolution data on surface geometry and texture, even under varying lighting \cite{Yuan2017, Dong2017}. From the original GelSight concept, a diverse range of VBTS designs have emerged to enhance performance and broaden application: GelSlim for more compact integrations \cite{Donlon2018}, GelSight Fin Ray for soft compliant grippers \cite{Liu2022}, and GelSight Svelte for human shape mimicking \cite{Zhao2023}, among others. Simulation tools have also evolved in parallel, accelerating the design and deployment of novel VBTS solutions \cite{Agarwal2021, Wang2022, Akinola2025, Zhong2025}.

Despite these advantages, conventional VBTSs face challenges for continuous large-area scanning. The rigid backing of the elastomer limits its ability to slide smoothly across surfaces, leading to distortions and potential damage to the sensing membrane. As a result, most operate in a discrete “press-and-lift” mode, which is slow for large-area inspection; for example, the commercial GelSight Mobile™ Series 2 requires approximately 0.1 seconds of exposure time per measurement patch \cite{GelSightMobile2024}, making large-area coverage time-consuming \cite{Agarwal2023}. Attempts at continuous sliding, such as with the DIGIT sensor for high-speed Braille reading \cite{potdar_high-speed_2024}, have shown potential but also revealed limitations: reliance on a conventional camera leads to motion blur at higher speeds, reducing recognition accuracy, while friction during sliding can cause rapid wear or damage to the elastomer. Protective layers can mitigate wear but may introduce optical diffusion, limiting fine texture resolution, and require precise contact control to ensure consistent data acquisition.

Neuromorphic vision sensors, or event cameras, offer a transformative solution to these speed and motion blur limitations \cite{Gallego2022}. Operating asynchronously at the pixel level, they detect brightness changes with microsecond temporal resolution, high dynamic range, low latency, and inherent suppression of motion blur \cite{Brandli2014, Posch2011, Lichtsteiner2008}. These properties are well-suited to capturing the rapid, fine-scale elastomer deformations that occur during high-speed tactile scanning. Several Neuromorphic VBTSs (NVBTSs) have leveraged these advantages: Gaussian processes on event data for contact force estimation \cite{Naeini2020}, spiking neural networks for texture classification (NeuroTac) \cite{Ward-Cherrier2020}, event-based marker tracking for slip detection (Evetac) \cite{Funk2024}, graph neural networks for contact angle prediction (Tactigraph) \cite{Sajwani2023}, \revisionTwo{spiking neural networks for robust Braille reading in noisy environments \cite{Xu_Braille_Neuromorphic}, contact status and force estimation (GelEvent \cite{GelEvent})}, and event-based image representations for proprioception and contact analysis \cite{Faris2023}. While these works validate the potential of NVBTSs, they have primarily focused on grasping-related tasks (slip, force, contact classification) or local texture analysis. Moreover, for these use cases, \revised{the algorithms used in combination with most NVBTS often result in decreased spatial resolution due to their reliance on tracking sparse markers}, which is detrimental for surface reconstruction and depth estimation \cite{Funk2024}. 

\subsection{Roller-Based and Large Surface Tactile Sensors}
Roller-based tactile sensors represent a promising avenue for achieving continuous tactile sensing, particularly for large surfaces where traditional planar sensors are inefficient. Early work explored cylindrical designs for specific detection tasks, such as the sensor developed by Shimonomura et al. \cite{Shimonomura2021} for detecting foreign bodies like food impurities during rolling contact. Similarly, studies in robotic in-hand manipulation have successfully utilized a roller grasper, often equipped with high-resolution tactile feedback, for tasks like object insertion and regrasping \cite{Lepert2023, Yuan2025}, demonstrating the utility of rolling contact in dynamic scenarios beyond simple surface mapping.

Transitioning towards surface property characterization, TouchRoller \cite{Cao2023} demonstrated the potential for mapping surface textures by continuously rolling a sensor element across a fabric sample. It achieved coverage of an 8cm $\times$ 11cm area in 10 seconds, yielding a 2D texture map. Notably, its operational speed was limited to approximately 11 mm/s; the authors noted that higher speeds resulted in motion blur within the conventional camera frames, hindering performance. This motion blur problem is particularly acute for VBTSs, which require close-up imaging to resolve fine surface details. Operating at such close proximity means that even moderate rolling speeds translate into high velocities of features across the image sensor plane; during the camera's finite exposure time, these features traverse a larger number of pixels, resulting in significant motion blur compared to imaging from a distance. Furthermore, TouchRoller did not perform 3D surface reconstruction. Addressing 3D reconstruction, GelRoller \cite{zhang_gelroller_2024} employed a cylindrical design integrating photometric stereo techniques for local surface normal estimation from single frames. For reconstructing larger surfaces, they proposed stitching multiple local reconstructions using Iterative Closest Point (ICP) registration. While demonstrating 3D capability, this work did not explicitly analyze or optimize the operational speed of the rolling process.

A fundamental challenge inherent to cylindrical roller designs, however, is that the tactile information gathered varies significantly with contact depth due to the sensor's curvature. This leads to inconsistencies and potential distortions in the acquired tactile images, complicating accurate reconstruction. Researchers have explored methods to mitigate this depth inconsistency. Li et al. \cite{Li2023} proposed a cyclic image fusion strategy, processing information from different contact phases within the rolling sequence using wavelet transforms to enhance information extraction and reduce bias from the curved contact. 

An alternative approach addresses the issue mechanically: the GelBelt sensor \cite{mirzaee_gelbelt_2025} utilizes a flat elastomeric belt moving between two wheels, creating a larger, more uniform contact area akin to a conveyor. This design avoids the depth variation inherent in direct cylindrical contact and enabled continuous sensing at higher reported speeds, up to 45 mm/s. \revisionTwo{Building on this hardware, GelSLAM \cite{GelSLAM} introduced a real-time tactile SLAM system capable of estimating object pose and reconstructing global shape by fusing local surface normals.} However, this design comes with some drawbacks. It relies on sliding motion to achieve high-speed operation, using an air gap and a smooth surface to reduce friction. While this allows GelBelt to move quickly, it creates challenges for lighting and can lead to wear over time, which may affect the system’s accuracy and repeatability. The design also struggles with curved surfaces. Its large, flat structure makes it hard to access concave areas, and it doesn’t perform well on convex surfaces either. In this case, the flat sensor doesn’t match the shape of the curved surface, leading to a limited sensing area, similar to problems seen in roller-based systems.

Furthermore, despite these advances in measurement continuity, including GelBelt's increased speed, systems relying on conventional cameras remain fundamentally constrained by the trade-off between speed and accuracy due to motion blur. As demonstrated by GelBelt's own results showing decreased accuracy at higher speeds \cite{mirzaee_gelbelt_2025}, pushing scanning velocities further significantly degrades the quality of the captured tactile information and subsequent reconstructions. While naively increasing camera frame rates might seem like a straightforward solution to combat motion blur, it introduces substantial challenges regarding the increased computational complexity required for real-time processing, significantly larger data storage requirements, and higher energy consumption, potentially limiting practical deployment. In turn, the potential for using event-based vision in roller sensors to achieve similar resolution at low computational power and significantly higher speeds for continuous tactile sensing is considerable. However, this capability has yet to be demonstrated, particularly for demanding large-scale inspection tasks.

\section{\revised{Background - Event Camera Principles}}
Unlike conventional cameras that capture intensity information across all pixels at fixed frame rates, event cameras employ a fundamentally distinct, bio-inspired sensing paradigm \cite{Gallego2022}. Each pixel operates independently and asynchronously, responding dynamically to changes in illumination. Specifically, a pixel monitors the logarithm of the incident light intensity, \( L(x, y, t) = \log(I(x, y, t)) \). An event \( e_k = (x_k, y_k, t_k, p_k) \) is generated at coordinates \( (x_k, y_k) \) only when the change in this logarithmic intensity, \( \Delta L \), relative to its value at the time of the last event from that pixel, surpasses a predefined contrast threshold, \( \pm C_\text{th} \). The generated event includes a high-resolution timestamp \( t_k \) and a polarity \( p_k \in \{-1, +1\} \), signifying the direction of the brightness change (decrease or increase, respectively). Following event generation, the pixel's internal reference brightness level is updated, priming it for subsequent intensity variations. The output is thus not a sequence of frames, but a sparse, continuous stream of asynchronous events reflecting dynamic scene changes. The unique characteristics of event cameras offer substantial benefits for high-speed tactile inspection. Their inherent high temporal resolution, with events timestamped often at microsecond precision, allows for tactile sensing at high speeds without the motion blur experienced by other roller tactile sensors at similar speeds. Furthermore, event-based cameras benefit from low latency and high dynamic range allowing for potential reduction in computational load and robustness to varying illumination within the sensor housing respectively.

\section{\revised{Methodology}}
\subsection{Mechanical Design and Fabrication}

\begin{figure}[thpb]
      \centering
        \includegraphics[width=\linewidth]{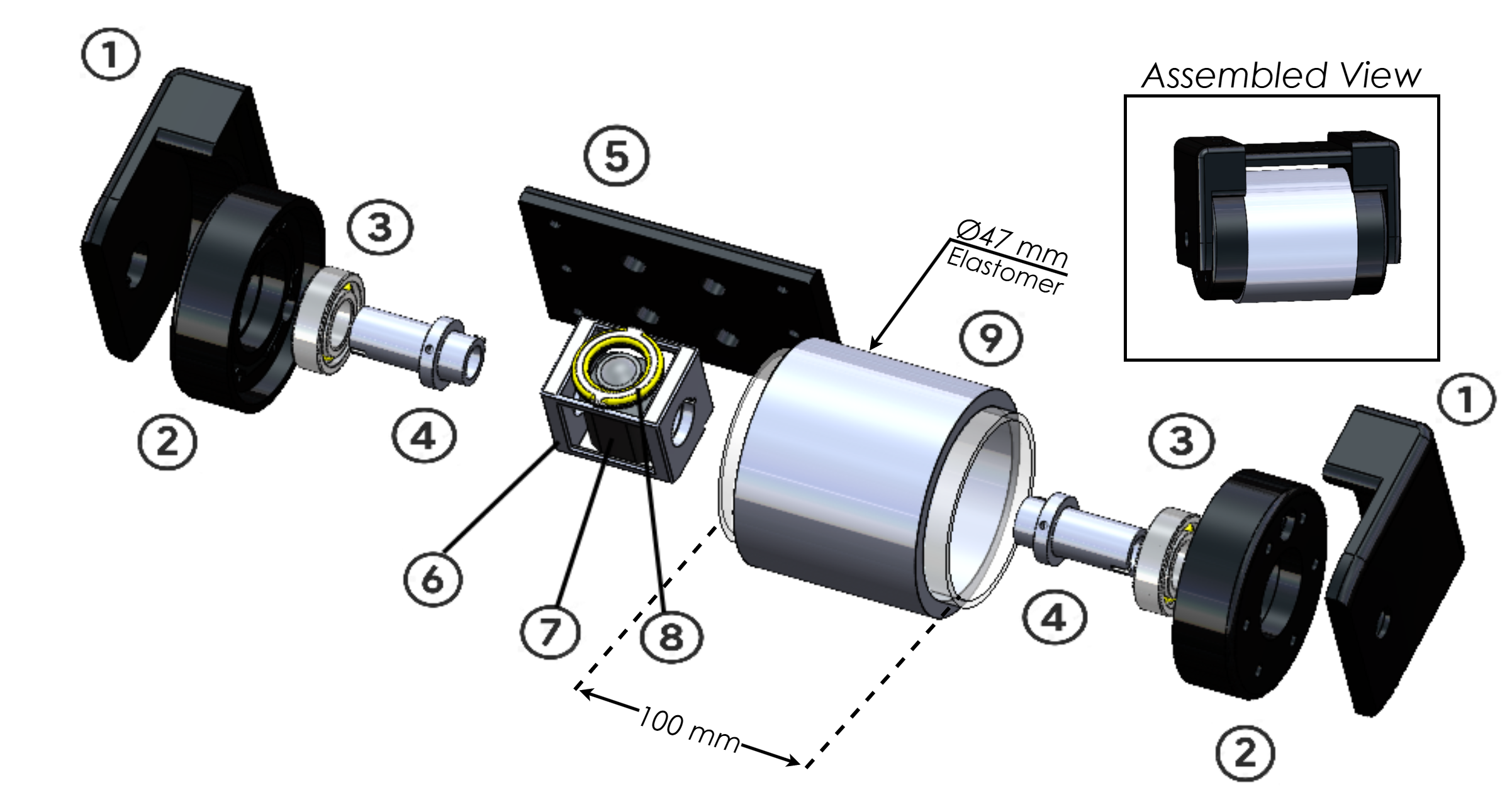}
      \small
      \circled{1} Side Connector\quad\circled{2} Side Cover\quad\circled{3} Bearing\quad\circled{4} Rotating Shaft\quad\circled{5} Top Connector\quad\circled{6} Camera Holder\quad\circled{7} DVXplorer mini camera\quad\circled{8} LED Ring with Diffuser\quad\circled{9} Acrylic Cylinder and Elastomer \\
      \caption{\revised{Mechanical Design of Roller: Exploded view of roller components with roller dimensions and inset of assembled view.}}
      \label{fig:mechanical_design}
\end{figure}

As illustrated in Figure \ref{fig:mechanical_design}, our proposed sensor consists of: a clear, transparent acrylic cylinder as a support structure; a translucent elastomer with a reflective coating; a neuromorphic camera (DVXplorer mini); a white LED ring with diffuser; shafts that attach to the camera holder; bearings to keep the camera holder stationary during rolling; and sealing covers to attach the cylinder and shafts to the outer housing as well as seal the light from the LEDs within the sensor.

The acrylic cylinder is cut to 100 mm in length, and the elastomer covers 80 mm of the cylinder. The elastomer layer coated with the reflective membrane covers a portion of the cylinder. When an object makes contact with the elastomer, the soft elastomer will deform. The camera will capture the deformation with the help of the reflective membrane and the illumination provided by the internal LEDs. Crucially, unlike photometric stereo methods which require precisely calibrated and stable illumination to infer shape from shading, our approach utilizes Event-Based Multi-View Stereo (EMVS). This method reconstructs 3D structure by tracking the apparent motion of features across multiple viewpoints as the sensor rolls (Structure from Motion). In turn, the system is inherently more robust to variations in illumination intensity or minor inconsistencies in LED positioning compared to the photometric stereo approach adopted by most VBTSs.

The camera is mounted and secured in the camera holder within the cylinder, with the entire sensor having dimensions of L: 116 mm, W: 148 mm, and H: 88 mm, with the elastomer being 47 mm in radius. The sensing area is around 40 by 30 mm.

The fabrication process begins with the elastomer. The elastomer is expected to be soft so that it is able to respond to contact even with a small force, and translucent so that the light can be reflected properly from the sensing area to the camera. For the fabrication of the elastomer, Smooth-On Sorta-Clear 18 with Shore A Hardness of 18 was used. Part A: Part B ratio was 10:1 as recommended by the manufacturer. A mold was 3D printed with two concentric acrylic cylinders to ensure a smooth outer surface for the elastomer and the cylinders were removed once curing was done.

Once the elastomer has cured and demolded, the reflective coating is applied consisting of: Inkcups Clear SI Ink, Hardener, Solvent, and silver mica powder in the ratios 8:1:1:1 respectively. The ink coating is then placed on the elastomer with a rotary silk screen printing machine and left to dry in an oven at 100°C for 10 mins as recommended by Inkcups to bond with the silicone elastomer.

The supporting parts, which were used to connect all the components, consisted of shafts, a camera holder, and sealing covers. The other components (shafts, camera holder, sealing covers) were 3D printed with PLA. Lastly, brass inserts and screws were used to secure the components together. \revised{The final assembled roller is shown (as an insert) in Figure \ref{fig:mechanical_design}}. \revised{The design files for the roller are available at: https://anonymous.4open.science/r/TSMR\_design\_files-6269/}

\subsection{Surface Reconstruction}
\subsubsection{\revised{Event-Based Multi-View Stereo (EMVS)}}
\label{subsec:EMVS}
In motion parallax, objects closer to the observer (or camera) appear to move faster across the field of view than objects farther away. This effect is due to the relative difference in angular displacement caused by motion, and it’s a key depth cue used by both biological and artificial vision systems to perceive 3D structure from motion. To estimate the 3D structure of the scanned surface from the event stream generated by the moving roller, we adapt the multi-view stereo (MVS) space-sweep method, originally developed for frame-based cameras \cite{collins_1996_a} and later extended to event cameras (EMVS) \cite{rebecq_2017_emvs}. This technique leverages the known viewpoints (poses) of the event camera over time. The core idea involves discretizing a volume of interest, \(\bar{D} \in \mathbb{R}^{W\times H\times N_z}\), encompassing the potential 3D scene, where \(W\), \(H\), and \(N_z\) represent the dimensions of the volume. Each captured event, \(e_k = (x_k,y_k,t_k,p_k)\), is back-projected through this volume as a 3D ray using the camera's calibrated intrinsic parameters and its known pose at time \(t_k\).
\revised{For our purposes, the physical design of the roller sensor provides a strong prior on the depth range at which events occur. We exploit this to optimize the discretization of the volume $\bar{D}$ for both accuracy and computational efficiency. The parameters $W$ and $H$ are set to the camera's resolution of 640x480 respectively. Furthermore, the depth dimension is constrained to a 4 mm range, from a minimum depth of 43 mm to a maximum depth of 47 mm, corresponding to the expected range of elastomer deformation from the sensor's 47 mm radius. This volume is discretized into $N_z =$ 500 depth planes, yielding a high depth resolution of 8 µm per plane. This constrained search space is critical for achieving both high-speed processing and high-precision reconstruction.}

A Disparity Space Image (DSI) is employed to accumulate evidence for surfaces within the volume. The DSI acts as a 3D histogram where each voxel counts the number of back-projected event rays passing through it, typically referenced to a specific viewpoint. Voxels corresponding to real surfaces in the scene are expected to lie at the intersection of rays from multiple viewpoints, resulting in high ray counts. Local maxima in the ray density within the DSI are then identified as potential 3D points belonging to the scanned surface \cite{rebecq_2017_emvs,ayyad_2023_neuromorphic}. Processing these maxima using Adaptive Gaussian Thresholding, yields a semi-dense depth map relative to the reference viewpoint.

\revised{Our implementation, in line with the original EMVS framework \cite{rebecq_2017_emvs}, conceptually associates a unique pose with every event. In our hardware setup (refer to Section \ref{sec:setup}), the sensor pose is derived from the UR10 robot's end-effector state, which is logged at 100 Hz. To obtain a pose for any event's specific microsecond-resolution timestamp, we perform linear interpolation between the two nearest pose readings. For computational efficiency, as is done in EMVS, events are grouped into small temporal packets (comprised of 256 events) and all events in a packet are assigned the interpolated pose of the median event's timestamp.}

\revised{The warping function relies on standard camera projection models.} Event cameras utilize the same optical principles as traditional cameras, allowing the use of the pinhole projection model to relate 3D world points to 2D image points. A 3D point \(\mathbf{P}_W = [X_W, Y_W, Z_W]^T\) in the world frame is projected onto the image plane as a 2D pixel \(\mathbf{p} = [u, v, 1]^T\) according to:

\begin{equation}
z \mathbf{p} = \mathbf{K} [ \mathbf{R} | \mathbf{t} ] \begin{bmatrix} \mathbf{P}_W \\ 1 \end{bmatrix} = 
\mathbf{K} \mathbf{P}_C
\label{eq:pinhole}
\end{equation}

where \(z\) is the depth \(Z_C\) of the point in the camera frame, \(\mathbf{K}\) is the \(3 \times 3\) camera intrinsic matrix, \([\mathbf{R} | \mathbf{t}]\) represents the \(3 \times 4\) extrinsic matrix (rotation \(\mathbf{R}\), \revised{which is a  \(3 \times 3\) matrix} and translation \(\mathbf{t}\), \revised{which is  \(3 \times 1\)}) transforming world coordinates to camera coordinates \(\mathbf{P}_C = [X_C, Y_C, Z_C]^T\). The absolute pose of the camera frame relative to the robot base frame, \({}^{B}\mathbf{T}_{C}(t)\), which defines \(\mathbf{R}\) and \(\mathbf{t}\) relative to the base frame at time \(t\), is obtained by combining the robot's end-effector pose \({}^{B}\mathbf{T}_{E}(t)\) (derived from forward kinematics) with the fixed hand-eye transformation \({}^{E}\mathbf{T}_{C}\), determined via hand-eye geometric calibration \cite{ayyad_2023_neuromorphic}. The intrinsic matrix \(\mathbf{K}\) is also obtained through calibration using E-Calib, a calibration toolbox for event cameras \cite{salah_2024_ecalib}.

The warping process transforms the depth value associated with a pixel \(\mathbf{p}_i = [u, v, 1]^T\) in the image plane at time \(t_i\) to the corresponding pixel location \(\mathbf{p}' = [u', v', 1]^T\) in the image plane at the common reference time \(t_m\). Specifically, this involves three steps:
\revised{
\begin{enumerate}
\item \textbf{Back-projection:} A pixel \(\mathbf{p}_i\) with depth \(Z_i(u,v)\) from the depth map at time \(t_i\) is back-projected to its corresponding 3D point \(\mathbf{P}_i = [X_i, Y_i, Z_i]^T\) in the camera's coordinate frame at \(t_i\) using the inverse pinhole projection:
\begin{equation}
\mathbf{P}_i = Z_i(u,v) \mathbf{K}^{-1} \mathbf{p}_i
\label{eq:back_projection}
\end{equation}
\item \textbf{Transformation:} The 3D point \(\mathbf{P}_i\) is transformed into the camera's coordinate frame at the reference time \(t_m\) using the relative pose transformation \({}^{m}\mathbf{T}_{i}\), which is calculated from the absolute poses as \({}^{m}\mathbf{T}_{i} = ({}^{B}\mathbf{T}_{C}(t_m))^{-1} \cdot {}^{B}\mathbf{T}_{C}(t_i)\). Using homogeneous coordinates:
\begin{equation}
\begin{bmatrix} \mathbf{P}_m \\ 1 \end{bmatrix} = {}^{m}\mathbf{T}{i} \begin{bmatrix} \mathbf{P}_i \\ 1 \end{bmatrix}
\label{eq:transformation}
\end{equation}
where \(\mathbf{P}_m = [X_m, Y_m, Z_m]^T\) is the point in the reference frame \(m\).
\item \textbf{Re-projection:} The transformed 3D point \(\mathbf{P}_m\) is projected back onto the image plane at time \(t_m\) to find the warped pixel coordinates \(\mathbf{p}' = [u', v', 1]^T\):
\begin{equation}
Z_m \mathbf{p}' = \mathbf{K} \mathbf{P}_m
\label{eq:re_projection}
\end{equation}
\end{enumerate}
}

\subsubsection{\revised{Multi-Reference Depth Map Generation and Fusion}}
The standard EMVS approach, described above, generates a depth map referenced to a single key viewpoint, often chosen as the midpoint \(t_m = (t_s+t_e)/2\) within a processing time window \([t_s, t_e]\). We select a time window of 20 ms, balancing the need for sufficient event integration for reconstruction accuracy with the requirement for high-speed operation (up to 0.5 m/s) without introducing significant motion artifacts in the DSI. \revised{This 20 ms time window refers to the total duration of event data used to generate a single depth map, and this window contains numerous event packets, each with a distinct pose, ensuring multi-view stereo reconstruction.}

However, roller-based tactile sensing can suffer from errors due to the sensor's curved nature, leading to inconsistent contact depth and potential fabrication variations \cite{li_2025_cyclic}. \revised{The curved elastomer means that even when contacting a flat surface, the contact points have varying depths relative to the camera. Second, the small contact area means any given surface feature is only observed over a short rolling distance within a short time window, resulting in a small baseline for multi-view stereo, which can make depth estimation more challenging.} High-speed motion can also introduce occlusion and warping artifacts. To enhance robustness against these effects and specifically mitigate the depth inconsistency and curvature-induced errors inherent in cylindrical tactile sensing, we extend the EMVS framework by generating and fusing depth maps from multiple reference times within the same time window. 

\revised{Inspired by works demonstrating the benefits of multi-reference objectives in event-based vision \cite{shiba_2024_secrets} and fusion techniques in roller tactile sensing \cite{li_2025_cyclic}, our approach extends the EMVS framework through a three-stage fusion process. First, we generate three separate DSIs using the same 20 ms stream of events and their associated poses. Each DSI is constructed relative to a different reference pose: one at the start (\(t_s\)), one at the midpoint (\(t_m\)), and one at the end (\(t_e\)) of the time window. Although the input data is identical, the differing reference frames result in distinct DSI geometries and consequently, slightly different ray count distributions. Second, as is done in \cite{rebecq_2017_emvs}, for each of the three DSIs, we extract a deterministic depth map by performing a maximization along each pixel's viewing ray to find the depth plane with the highest vote count. This yields three independent depth maps: \(Z_s\), \(Z_m\), and \(Z_e\) referenced at the start ($t_s$), midpoint ($t_m$), and end ($t_e$) times of the window, respectively.}

\begin{figure}[thpb]
    \centering
        \includegraphics[width=\linewidth]{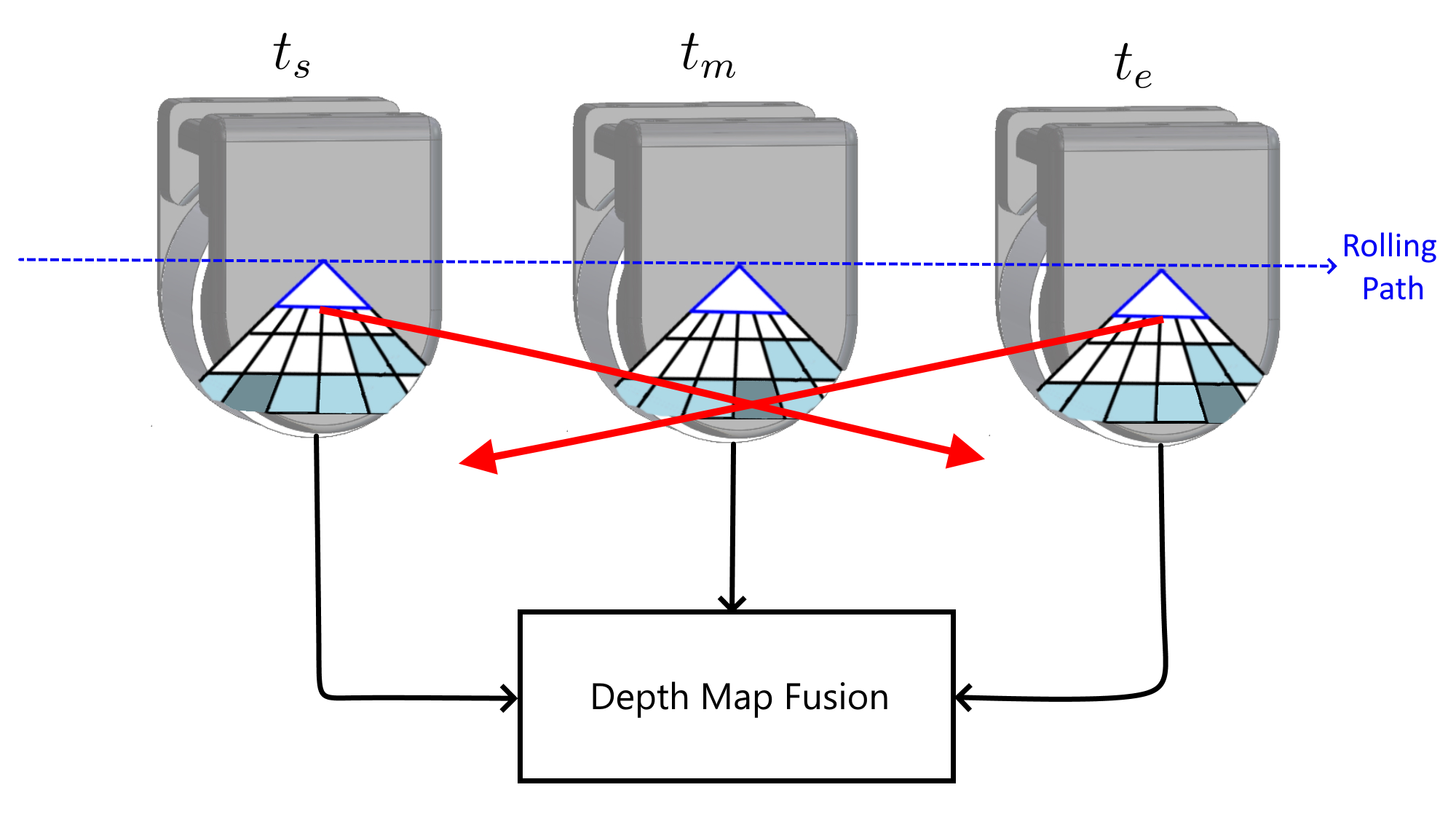}
    \caption{\revised{Multi-Reference Fusion Strategy: The roller sensor captures depth maps at three reference times within the time window: start (\(t_s\)), midpoint (\(t_m\)), and end (\(t_e\)). The blue dashed horizontal line indicates the rolling path of the sensor. The light blue triangular frustums represent the DSI ray counters for each reference time. The red arrows represent incoming viewing rays from back-projected events, which vote for all the DSI voxels they pass through. Depth maps are extracted from each DSI and fused using Bayesian Model Averaging (BMA).}}
    \label{fig:DSI_roller}
\end{figure}

\revised{Third, we fuse these individual depth maps using a Bayesian Model Averaging (BMA) strategy (Figure \ref{fig:DSI_roller}), which has proven effective for combining multiple noisy estimates in vision tasks \cite{sun_2018_bayesian}. This process averages out noise and inconsistencies present in the individual reconstructions, mitigating both DSI discretization artifacts and the impact of noise and occlusions. The final fused depth map, \(Z_f\), is computed as a weighted average of the three deterministic maps after warping them to a common reference frame (chosen as the midpoint view, \(t_m\)):}

\begin{equation}
    Z_{f} = \sum_{i \in \{s,m,e\}} w_i \cdot W_i(Z_i)
    \label{eq:fusion}
\end{equation}

where \(Z_i\) is the depth map generated at reference time \(t_i\), \(w_i\) is the corresponding fusion weight, \revised{and \(W_i\) is the warping function described in Section \ref{subsec:EMVS}. The original depth value \(Z_i(u,v)\) is associated with the calculated warped pixel location \((u', v')\) in the reference frame \(m\) before the weighted averaging is performed.}

\subsection{Parameter Calibration}
\label{sec:calibration}

To optimize the final reconstruction quality, we perform a one-time calibration using an object with precisely known geometry, specifically a sphere of known radius \(R_\text{sphere}\). This calibration determines the optimal values for both the adaptive thresholding parameters used in processing the DSI and the fusion weights used in combining multi-reference depth maps. The adaptive thresholding parameters (constant $C$ and median filter size $G_f$) control the density and noise level of the individual depth maps \(Z_i\), while the fusion weights \(w_s, w_m, w_e\) dictate how these maps are combined into the final estimate \(Z_f\). Furthermore, this calibration procedure can be readily repeated in-situ, enabling the system to adapt to gradual changes in the elastomer surface, such as wear, by re-optimizing these parameters.

\begin{figure}[thpb] 
    \centering
    \includegraphics[width=0.8\linewidth]{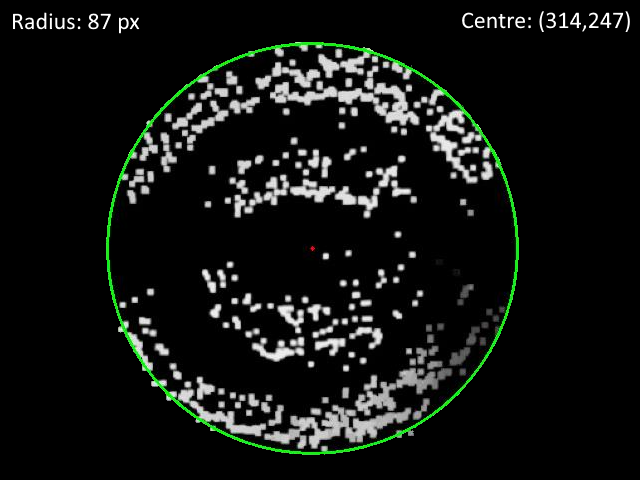}
    \caption{Identification of the circular contact patch from depth map during calibration with the 4mm radius sphere. The green circle outlines the detected boundary, from which the contact radius \(r\) and center \((x_b, y_b)\) (red dot) are determined in pixel coordinates.}
    \label{fig:sphere_contact_patch}
\end{figure}

To establish the ground truth depth map \(Z_\text{gt}\) for the sphere contact, we adapt the methodology used in \cite{zhang_gelroller_2024}. \revised{For calibration, the sensor is rolled over the calibration sphere while recording event data. The sphere is mounted in a fixed position on an optical breadboard such that the sphere passes directly under the camera’s field of view during rolling, ensuring consistent alignment (refer to setup in Figure \ref{fig:setup})}. Then, the resulting circular contact patch in the depth map is identified (visualized in Figure \ref{fig:sphere_contact_patch}). From this, we determine the radius of the pressed circle, \(r\) (in pixels), and the coordinates of its center, \((x_b, y_b)\) (in pixels). The physical radius of the calibration sphere, \(R_\text{sphere}\), is known in millimeters. \revised{The measured contact radius, \(r\), (in pixels) is then converted to millimeters by multiplying by the scale factor \(f\) (mm/pixel), where \(f = 0.0422\) mm/pixel. This scale factor was determined by imaging the gap between the jaws of a Vernier caliper indented directly on the sensor surface at known separations. Two readings were taken at each of four separations (2 mm, 3 mm, 4 mm, and 5 mm), the scale factor was computed for each reading, and the eight resulting values were averaged to obtain the final scale factor \(f\). The depth of the sphere's center, \(z_b\) (in mm), can then be calculated geometrically as:
\begin{equation}
z_b = -\sqrt{R_\text{sphere}^2 - (r \cdot f)^2}
\end{equation}
For any pixel \((x_i, y_i)\) within the identified contact circle, its depth \(z_i\) (in mm) must satisfy the sphere equation:
\begin{equation}
\big((x_i - x_b) \cdot f\big)^2 + \big((y_i - y_b) \cdot f\big)^2 + (z_i - z_b)^2 = R_\text{sphere}^2
\label{eq:sphere_calib}
\end{equation}
Here, \((x_i - x_b)\) and \((y_i - y_b)\) are multiplied by the scale factor \(f\) to convert from pixels to millimeters before solving the equation. Solving Equation \ref{eq:sphere_calib} for \(z_i\) for all pixels inside the contact circle yields the ground truth depth map directly in millimeters.}

With the ground truth \(Z_\text{gt}\) established, both sets of parameters (thresholding and fusion weights) are optimized simultaneously by minimizing the Mean Absolute Error (MAE) between the final fused depth map \(Z_f\) and \(Z_\text{gt}\). A region mask corresponding to the sphere's contact area is applied before calculating the MAE. The optimization is performed via a grid search over the combined parameter space:
\begin{equation}
    \min_{\theta} \, \mathbb{E}[||Z_{f} - Z_\text{gt}||_1] 
    \label{eq:combined_opt}
\end{equation}
where $\theta = \{C,G_f, w_s, w_m, w_e\}$. The range and optimal values for each parameter can be seen in Table \ref{parameters}.

\begin{table}[h]
\caption{Parameter Ranges and Optimal Calibrated Values}
\label{parameters}
\begin{center}
{
\resizebox{0.85\linewidth}{!}{
\begin{tabular}{|c|c|c|}
\hhline{===}
\multicolumn{1}{c}{Parameter} & \multicolumn{1}{c}{Range} & \multicolumn{1}{c}{Optimal Value}\\
\hhline{===}
$C$ & 5--20 & 16 \\
\hline
$G_f$ & 11--91 & 45 \\
\hline
$w_m$ & 0--1 & 0.333 \\
\hline
$w_s$ & 0--1 & 0.445 \\
\hline
$w_e$ & 0--1 & 0.222 \\
\hline

\end{tabular}
}
}
\end{center}
\end{table}

\begin{figure}[thpb]
    \centering
    \includegraphics[width=\linewidth]{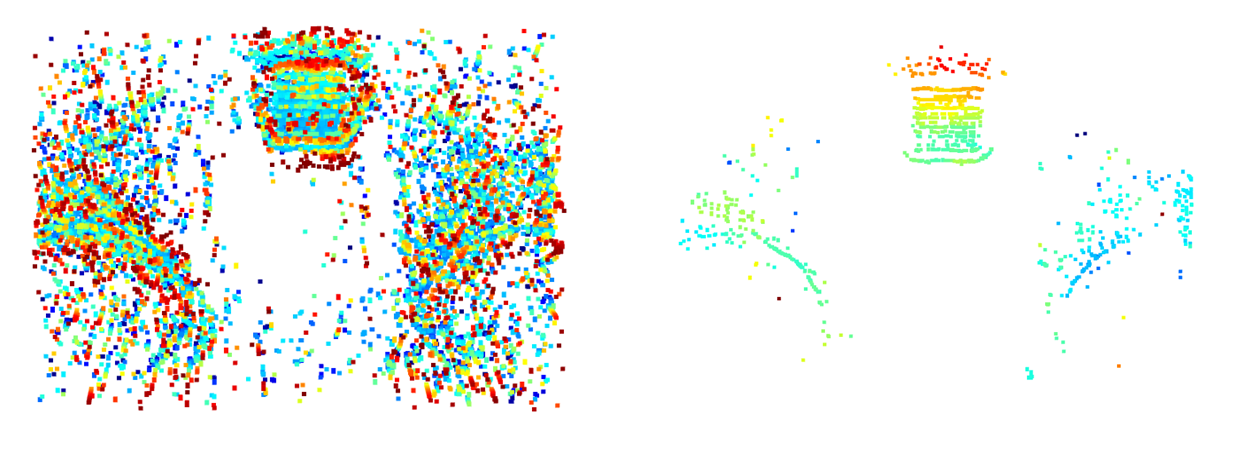} 
    \caption{Comparison of pointcloud processed with EMVS: before calibration (left) and after calibration (right). The pointcloud processed with calibrated parameters has significantly less noisy events particularly around the tactile object of interest. Color represents change in depth using JET colormap.}
    \label{fig:calibration_effect}
\end{figure}

The calibrated values for all parameters are subsequently fixed and utilized for processing the data from all other scanned objects presented in this work. The effect of calibration on the processed depth map and point cloud is shown in Figure \ref{fig:calibration_effect}. From a Bayesian perspective, the optimal weights should ideally be inversely proportional to the uncertainty associated with each individual depth map \(Z_i\), effectively giving more influence to more confident estimates.

\subsection{\revised{Braille Feature Recognition}}
\label{sec:braille}
\revised{To demonstrate the sensor's capability for fine feature detection at high speed, a task complementary to 3D geometric reconstruction, we implemented a Braille character recognition pipeline. This task serves as a well-established benchmark in the tactile sensing community for evaluating high-speed spatial resolution, allowing for a direct comparison with state-of-the-art systems like that of Potdar et al. \cite{potdar_high-speed_2024}. This method specifically evaluates the sensor's ability to capture fine textural details directly from the event stream, a key aspect of tactile inspection that is distinct from geometric reconstruction.} The process begins by applying motion compensation \cite{Salah2023, Gallego_CMAX_2018} to the recorded event stream as the sensor rolls over the Braille plate. Given the constant linear velocity of the roller, \(v_{\text{roller}}\), which is commanded to and precisely controlled by the UR10 robotic arm as described in the Experimental Setup (Sec. \ref{sec:setup}), we can determine the corresponding optical flow \( \mathbf{\dot{U}} = (\dot{u}, \dot{v}) \) of features moving across the image plane with $[u,v]$ as the pixel coordinates. The roller's motion involves predominantly horizontal motion, therefore \( \dot{v} \approx 0 \) and \( \dot{u} = \frac{-F}{Z} v_{\text{roller}} \), where $F$ is the focal length of the camera and $Z$ is the depth of the scene (assumed to be constant depth of 47 mm which is the radius of the roller with maximum deformation of 4.5 mm).

Without motion compensation, if we were to simply create a frame-like image by plotting all events recorded within a specific time window at their original pixel locations (a process called direct accumulation), the relative movement between the sensor and the Braille plate would cause significant motion blur. This occurs because events generated by a single point on the moving Braille surface would be spread across multiple pixels in the resulting image (as seen in Figure \hyperref[fig:braille_pipeline]{\ref*{fig:braille_pipeline}a}). Using a long accumulation time integrates more events but exacerbates this blur as features traverse many pixels, whereas a very short accumulation time minimizes blur but often captures too few events to form a clear representation, leading to degraded features \cite{Salah2023, Gallego_CMAX_2018}. To counteract this, we warp events with batches consisting of a constant number of events ($N_e$ = 20,000 events) at the initial timestamp of the event stream as the reference time, $t_{\text{ref}}$. For each event \( e_k = (\mathbf{x}_k, t_k, p_k) \) where \( \mathbf{x}_k = (x_k, y_k) \), its warped position \( \mathbf{x}'_k \) is calculated as:
\begin{equation}
    \mathbf{x}'_k = \mathbf{x}_k - \mathbf{\dot{U}} (t_k - t_{\text{ref}})
    \label{eq:event_warp}
\end{equation}
A motion-compensated frame, or Image of Warped Events (IWE) \(I_{\text{warp}}\), is generated by accumulating the counts of warped events into a 2D histogram (ignoring the polarities of each event):
\begin{equation}
    I_{\text{warp}}= \sum^{N_e}_{k} \delta(\mathbf{x} - \mathbf{x}'_k)
    \label{eq:iwe_count}
\end{equation}
where \( \delta(\cdot) \) is the Dirac delta function and $\mathbf{x}$ represents a pixel in the IWE. This process effectively integrates information over the batch into an image frame at any speed, creating a sharper representation of the Braille dots.

\section{\revised{Experimental Setup And Procedures}}
\label{sec:setup}

\subsection{\revised{Experimental Setup}}
\begin{figure}[thpb]
    \centering
    \includegraphics[width=\linewidth]{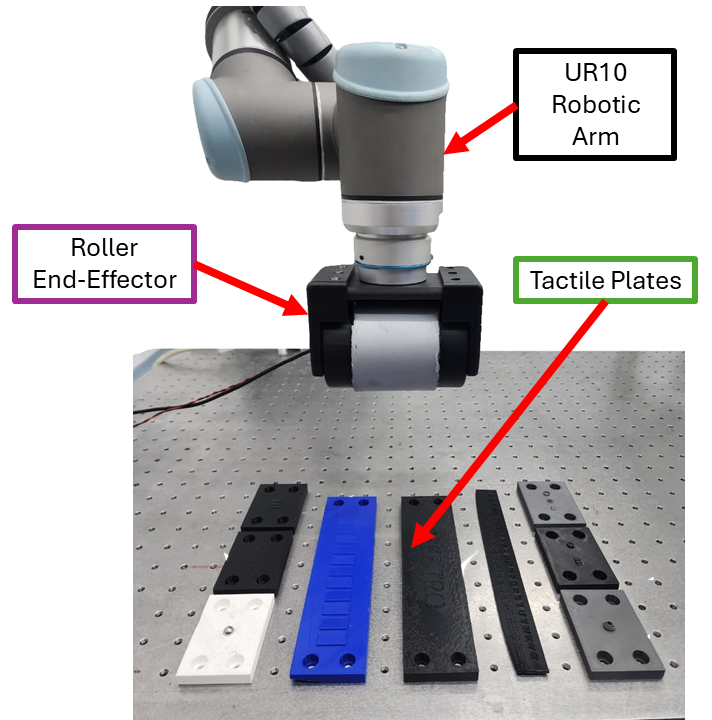}
    \caption{Experimental setup showing the neuromorphic roller sensor mounted as an end-effector on a UR10 robotic arm. The robot controls the rolling motion over the mounted tactile plates.}
    \label{fig:setup}
\end{figure}

To evaluate the performance of the proposed neuromorphic roller tactile sensor, a series of experiments were conducted focusing on its capabilities in 3D surface reconstruction and tactile feature recognition under controlled conditions. The roller sensor was mounted as an end-effector onto a Universal Robots UR10 robotic arm, as depicted in Figure \ref{fig:setup}. This setup provided precise control over the linear rolling trajectory and speed of the sensor across various target surfaces, which is crucial for systematic evaluation and understanding the effect of operational parameters. The end effector velocity was recorded directly from the UR10 robot's real-time data interface and logged during the experiments. Experiments were conducted at various controlled speeds, ranging from 0.05 m/s to 0.5 m/s, allowing for analysis of the sensor's performance under different dynamic conditions.

A variety of objects and surfaces were utilized for testing:
\begin{itemize}
    \item A calibrated metal sphere (4 mm radius) was used for parameter calibration and quantitative accuracy assessment of single-object reconstruction.
    \item \revised{Several 3D printed PLA objects with diverse geometries: stairs with fine gradations, sparse sphere (concentric circles), hexagon with a shallow recess, bumps or semi-cylinders side by side with varying radii, and a pyramid (see Fig. \ref{fig:single_recon}) were used to test the reconstruction capabilities on varied shapes.}
    \item \revised{Two larger 3D printed plates were used to evaluate performance in large surface reconstruction tasks involving stitching multiple frames (see Fig. \ref{fig:surface_recon}): a line array with varying heights and a letter plate with varying-height letters and induced defects (e.g., scratches and printing artifacts).}
    \item Standard Braille plates (Fig. \ref{fig:a_z_braille}) were used to assess the sensor's ability for high-speed fine feature detection and recognition.
\end{itemize}

The subsequent Results section (Sec. \ref{sec:results}) details the outcomes of these experiments. The evaluations focus on: (1) 3D reconstruction accuracy, primarily quantified using Mean Absolute Error (MAE) against ground truth geometries, including an analysis of the proposed BMA fusion strategy; (2) the relationship between rolling speed and reconstruction accuracy; (3) the effectiveness of large surface reconstruction through measurement stitching; and (4) the speed and accuracy achieved during high-speed Braille character recognition. \revisionTwo{To quantify reconstruction accuracy, we calculate the Mean Absolute Error (MAE) between the reconstructed depth map, $Z_f$, and the ground truth, $Z_\text{gt}$. Due to the stochastic and sparse nature of event generation \cite{stochas}, a fixed dense ground truth grid cannot be used. Instead, we perform a sparse pixel-wise evaluation: for every reconstructed point at coordinates $(x, y)$ falling within the manually segmented contact region mask, the ideal depth $Z_\text{gt}(x, y)$ is computed analytically using the known object geometry. This ensures that $Z_\text{gt}$ is inherently pixel-wise aligned with $Z_f$, isolating depth estimation error from planar motion artifacts. The MAE is then computed as the average absolute difference for all valid sparse points within the mask. Notably, noise within the contact region is included in the error metric, while off-contact background noise (outside the contact region mask is excluded.} All reported MAE values are computed only on the sparse measurement points output by our method. Pixels without a valid depth estimate (shown as white in the visualized point clouds in Section \ref{sec:results}) are excluded from the error computation.

\subsection{\revised{Calibration Cross-Validation and Evaluation}}
\label{subsec:calibration_evaluation}
A potential concern from the calibration performed, as described in Section \ref{sec:calibration}, is that the optimal fusion weights determined using only the sphere geometry might be biased and not generalize well to other object shapes (outlined in detail in Section \ref{sec:setup}). To investigate this, we performed a cross-validation analysis. \revised{For calibration, the sensor is rolled over the surface of the calibration sphere/object (rather than pressing the sphere into a static roller as is done typically with VBTS/NVBTS) while recording event data. From each rolling pass, a point cloud is generated and the optimization in Eq. \ref{eq:combined_opt} is performed within the masked contact region to determine the optimal parameters.} We first determined the object-specific optimal fusion parameters \(\theta_i\) (the weights \(w_s, w_m, w_e\) were highlighted since the thresholding parameters did not change across objects) by running the grid search optimization (Eq. \ref{eq:combined_opt}) using the ground truth \(Z_\text{gt}^{(i)}\) for each object \(i\) (where \(i \in \{\text{Sphere, Stairs, Sparse Sphere}\}\)).

We then evaluated how well each set of parameters \(\theta_i\) performed on average across all \(M=3\) test objects. \revised{For each object \(i\), the calibration procedure was repeated for five independent rolling trials to ensure consistency of the estimated parameters. The MAE for object \(i\) was first computed for each trial and then averaged over the five trials to obtain a single representative MAE value for that object. Using these per-object averages, we then calculated the overall average MAE, \(e_i\), for parameter set \(\theta_i\) across all \(M=3\) objects as follows}:
\begin{equation}
e_i = \frac{1}{M}\sum_{i=1}^M\bigl\|Z_f^{(i)}(\theta_i) - Z_\text{gt}^{(i)}\bigr\|_1
\label{eq:cross_val_error}
\end{equation}
where \(Z_f^{(j)}(\theta_i)\) is the fused depth map generated for object \(i\) using parameters \(\theta_i\). \revised{The values reported in Table \ref{tab:cross_validation_weights} are therefore the result of a two-stage averaging process: first averaging the MAE over the five trials for each object, and then averaging these per-object MAEs across the three objects for each parameter set}. We then compared these error values to determine which calibration object yielded the parameters with the best overall generalization performance (i.e., the minimum \(e_i\)).

The results are summarized in Table \ref{tab:cross_validation_weights}. While the object-specific optimal weights vary slightly, the weights calibrated using the sphere yielded the lowest overall average MAE (\(e_{\text{Sphere}} = 0.0549\) mm) when applied to the objects within the calibration set, confirming its suitability as a general calibration target.

\begin{table}[htbp] 
\centering
\caption{Cross-validation of BMA fusion weights}
\label{tab:cross_validation_weights}
\resizebox{\linewidth}{!}{
\large
\begin{tabular}{|l|c|c|c|c|} 
\hhline{=====} 
\textbf{Calibrated on} & \textbf{\(w_s\)} & \textbf{\(w_m\)} & \textbf{\(w_e\)} & \revised{Avg. MAE of all objects (mm)} \\
\hhline{=====} 
Sphere & 0.445 & 0.333 & 0.222 & \textbf{0.0549} \\ \hline
Stairs & 0.5 & 0.333 & 0.167 & 0.0652 \\ \hline
Sparse Sphere & 0.4 & 0.4 & 0.2 & 0.0649 \\ \hline 
\end{tabular}
}
\end{table}

A consistent and perhaps counter-intuitive trend observed across all calibrations is that the start reference view (\(t_s\)) receives the highest (or tied for highest) weight, while the end view (\(t_e\)) consistently receives the lowest. Intuitively, one might expect the midpoint view (\(t_m\)), representing a temporal average, to be most reliable. However, this weighting suggests the information captured early in the time window is most critical for accurate reconstruction in our setup. \revised{The possible reasons for this trend and its implications are discussed in more detail in Section \ref{sec:discussion}.}

\subsection{\revised{Contact Force Analysis}}
\begin{figure}[h]
\centering
\includegraphics[width=\linewidth]{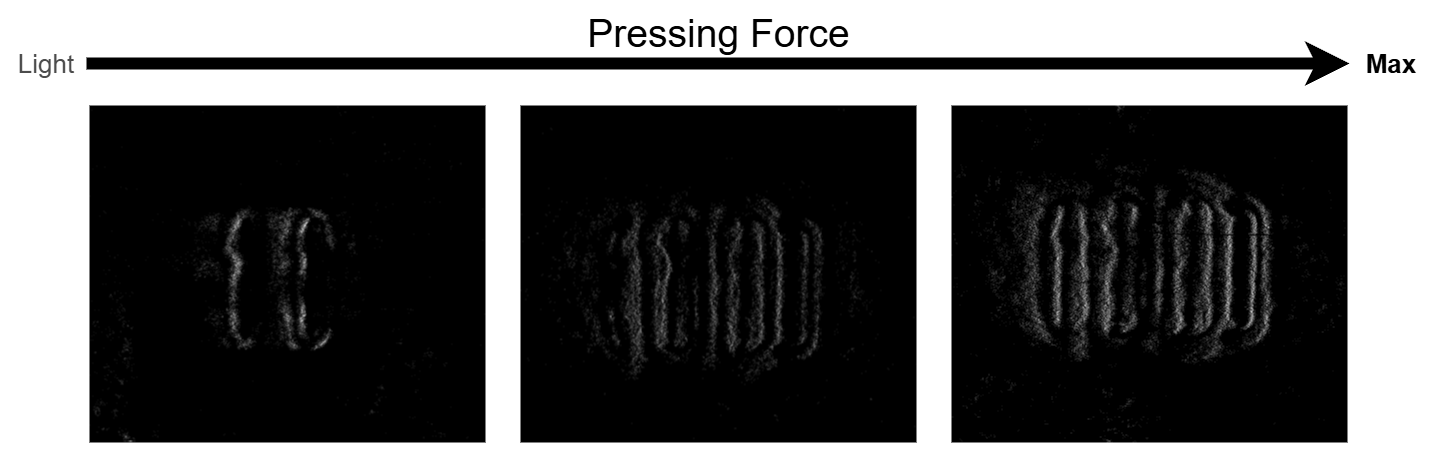}
\caption{\revised{Qualitative analysis of the effect of pressing force on signal quality using the bumps object. Motion-compensated event frames generated under light, medium, and maximum contact forces (left to right).}}
\label{fig:force_analysis}
\end{figure}

\revised{To evaluate the effect of normal contact force on reconstruction quality, we conducted a qualitative analysis. Using the UR10 robot's precise position control, we varied the vertical distance between the roller and the tactile plate to apply three distinct levels of elastomer deformation, corresponding to light, medium, and maximum contact forces. Figure \ref{fig:force_analysis} illustrates the resulting motion-compensated event frames generated from the bumps object under these conditions.}

\revised{The results show a clear relationship between contact force and signal quality. A light pressing force results in minimal deformation with the elastomer barely touching the plate, leading to weak event generation where critical features of the object such as the smaller semi-cylinders are not captured. In contrast, a maximum pressing force ensures the elastomer fully conforms to the surface geometry. This produces a strong, clear signal with high event counts, capturing the full geometry of the object and thus enabling the most accurate reconstruction. Based on this analysis, all results presented in this paper were obtained using a maximum contact force setting to ensure consistency and optimal data quality.}

\section{Results}
\label{sec:results}





\subsection{Single Object Reconstruction}
As seen in Table \ref{MAE_1}, the BMA fusion framework outperforms EMVS with an average percentage decrease in MAE of \textbf{25.2\%} across objects tested within the calibration set. Additionally, the fusion strategy of only fusing the left and right images that performed the best in reconstruction variability with Li et al. \cite{li_2025_cyclic} was tested. While their method performs better than using the depth map generated using baseline EMVS, the proposed BMA fusion framework still outperforms the other methods. \revised{A quantitative analysis on a line array (three identical rectangles with height 0.6 mm) further confirms the improvement in depth consistency: the standard deviation of the reconstructed depth across the line array decreased significantly from 0.067 mm for the baseline EMVS reconstruction to 0.028 mm when using the BMA fusion approach, indicating a much more uniform depth estimation.}

\begin{table}[h]
\caption{Evaluation of the tactile 3D reconstruction Average MAE in mm using different depth map fusion weights at 0.3 m/s with calibration set}
\label{MAE_1}
\begin{center}
{
\resizebox{\linewidth}{!}{
\large
\begin{tabular}{|c|c|c|c|c|}
\hhline{=====}
\multicolumn{1}{c}{Reconstruction Method} & \multicolumn{1}{c}{Stairs \revised{(mm)}} & \multicolumn{1}{c}{Sphere \revised{(mm)}} & \multicolumn{1}{c}{Sparse Sphere \revised{(mm)}} & \multicolumn{1}{c}{Average \revised{(mm)}}\\
\hhline{=====}
EMVS & 0.0672 & 0.1256 & 0.0274 & 0.0734 \\
\hline
Left \& Right Fusion & 0.0613 & 0.1085 & 0.0144 & 0.0614 \\
\hline
\textbf{BMA Fusion} &  \textbf{0.0581} & \textbf{0.0946} & \textbf{0.0120} & \textbf{0.0549} \\
\hline

\end{tabular}
}
}
\end{center}
\end{table}

\begin{figure}[h]
\centering
\includegraphics[width=\linewidth]{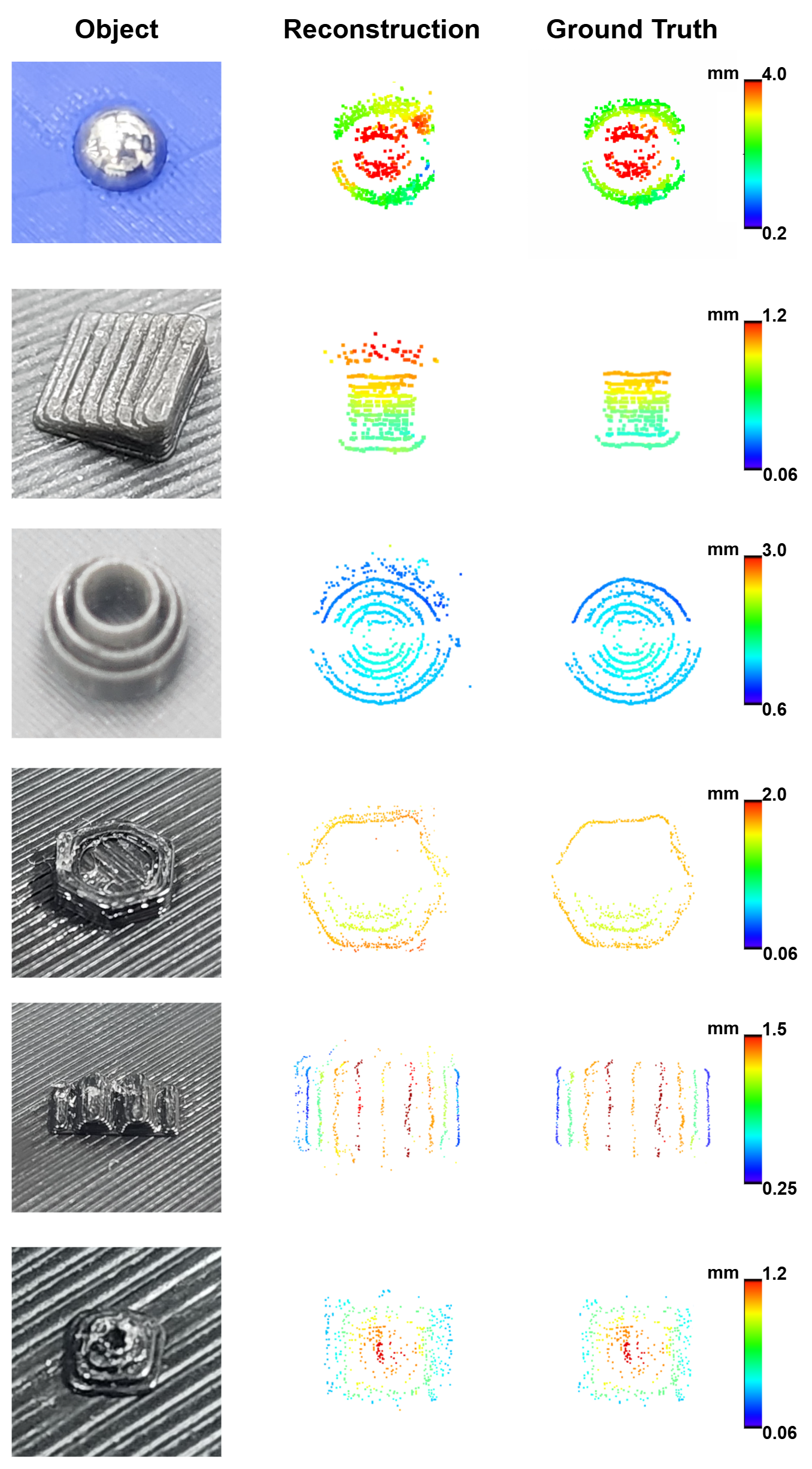}
\caption{\revised{Single object reconstruction. Objects from top to bottom: 4mm sphere, stairs, sparse sphere, hexagon with recess, bumps, pyramid.}}
\label{fig:single_recon}
\end{figure}

\revised{The comprehensive reconstruction accuracy for all six tested objects, each averaged over five independent trials, is presented in Table \ref{tab:all_objects_mae}. An analysis of these results reveals a consistent performance trend that correlates strongly with the operating principle of event cameras, which primarily generate signals at intensity edges during relative motion under
consistent illumination. The sparse sphere yielded the lowest MAE, as its distinct, well-separated edges with significant depth variance facilitate robust event generation. The stairs and pyramid objects, both possess well-defined, repeating features, but with smaller depth variations and in close proximity. These features could have potentially led to denser but less distinct event clusters and increased reconstruction error compared to the sparse sphere, although they were also reconstructed with high accuracy. The bumps object presents a hybrid case; the heights of the semi-cylinders and the sections between them were captured accurately, but their smooth, curved tops were not reconstructed. This is consistent with the limitations of the event-based algorithm on curved surfaces that lack edges or sharp features, similar to the solid sphere. The hexagon and sphere objects exhibited the highest MAE respectively. With regards to the hexagon, as seen in Figure \ref{fig:single_recon}, the reconstruction of the hexagon is incomplete, failing to capture the entire central recess. This can be attributed to self-occlusion, where the steep walls of the deformed elastomer block the camera's line of sight or perhaps the elastomer's Shore Hardness was not suitable to fully conform to the recess and capture its features accurately, leading to higher MAE overall. The solid sphere, possessing a smooth, continuous geometry without sharp edges (aside from the contact boundary), generates fewer internal feature-related events compared to the other objects (Fig. \ref{fig:single_recon}), resulting in a sparser representation and consequently the highest reconstruction MAE among the tested shapes.}

\begin{figure}[h]
    \centering
    \includegraphics[width=\linewidth]{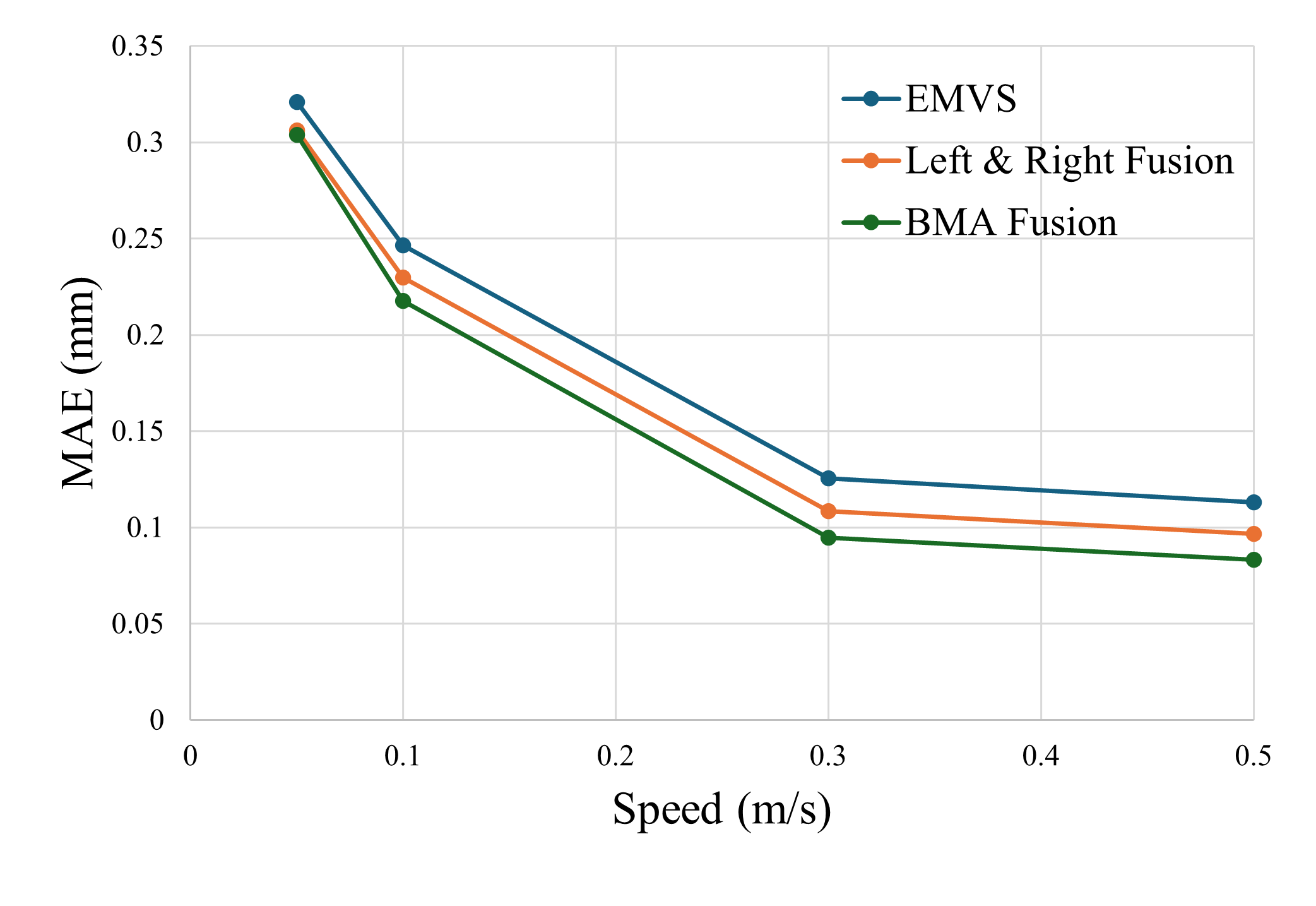} 

    \caption{Speed (m/s) against MAE (mm) for calibrated sphere comparing EMVS and BMA. MAE reduces as speed increases at exponentially decreasing rate.}
    \label{fig:speed_mae}
\end{figure}

BMA also generated depth maps with lower MAE across varying speeds as illustrated in Figure \ref{fig:speed_mae} in comparison to EMVS and Left \& Right Fusion. However, the reduction in the MAE of the sphere is more significant at higher speeds, with a 5.27\% reduction at 0.05 m/s, a 25.2\% reduction at 0.3 m/s and starts to plateau with a 26.4\% reduction at 0.5 m/s in comparison to EMVS. Interestingly, Figure \ref{fig:speed_mae} also shows that at lower speeds (e.g., 0.05 m/s), the performance difference between the BMA fusion (using all three reference views) and the Left \& Right fusion (using only start and end views) becomes negligible. This suggests that while BMA fusion provides substantial benefits at higher speeds by effectively combining information from more distinct viewpoints to mitigate motion artifacts and improve consistency, at lower speeds the depth maps generated at the start, middle, and end times are likely very similar due to the minimal camera displacement within the time window. Consequently, fusing only the start and end views captures most of the available information, and adding the highly correlated midpoint view offers little additional advantage. These findings are in line with Bhattacharya et al. \cite{bhattacharya2024monocular} as their event-based depth estimation model exhibits greater depth prediction quality at higher speeds. This effect could be explained by the fact that more events are generated within the same time window at higher speeds allowing for more accurate depth estimation through EMVS.

\revised{Another important consideration when it comes to BMA is the optimal number of reference views for fusion. To investigate this, we conducted an additional experiment on the sphere object using five reference views. This approach yielded a higher MAE of 0.1115 ± 0.0065 mm compared to the 0.0946 ± 0.0081 mm from the three-reference fusion, albeit with lower variability. This suggests a trade-off where adding more views can increase robustness but may also introduce additional warping errors that degrade overall accuracy. This finding is consistent with observations by Li et al. \cite{li_2025_cyclic}, who noted that excessive fusion can weaken unique information from individual viewpoints. The optimal number of reference views is likely task-dependent and differs depending on speed used, but for the conditions in our experiments, three views provided the best balance of accuracy and consistency.}

\begin{table}[h]
\caption{\revised{Reconstruction Accuracy for All Single Objects}}
\label{tab:all_objects_mae}
\begin{center}
{
\resizebox{\linewidth}{!}{
\large
\begin{tabular}{|c|c|c|}

\hhline{===}
\textbf{Object} & \textbf{Average MAE (mm)} & \textbf{Std. Dev. (mm)} \\
\hhline{===}
Sparse Sphere & 0.0120 & 0.0010 \\
\hline
Stairs & 0.0581 & 0.0026 \\
\hline
Bumps & 0.0589 & 0.0024 \\
\hline
Pyramid & 0.0677 & 0.0058 \\
\hline
Hexagon & 0.0808 & 0.0042 \\
\hline
Sphere & 0.0946 & 0.0081 \\
\hhline{---}
\textbf{Overall} & \textbf{0.0620} & \textbf{0.0047} \\
\hhline{===}
\end{tabular}
}
}
\end{center}
\end{table}

\begin{figure}[h]
\centering
\includegraphics[width=\linewidth]{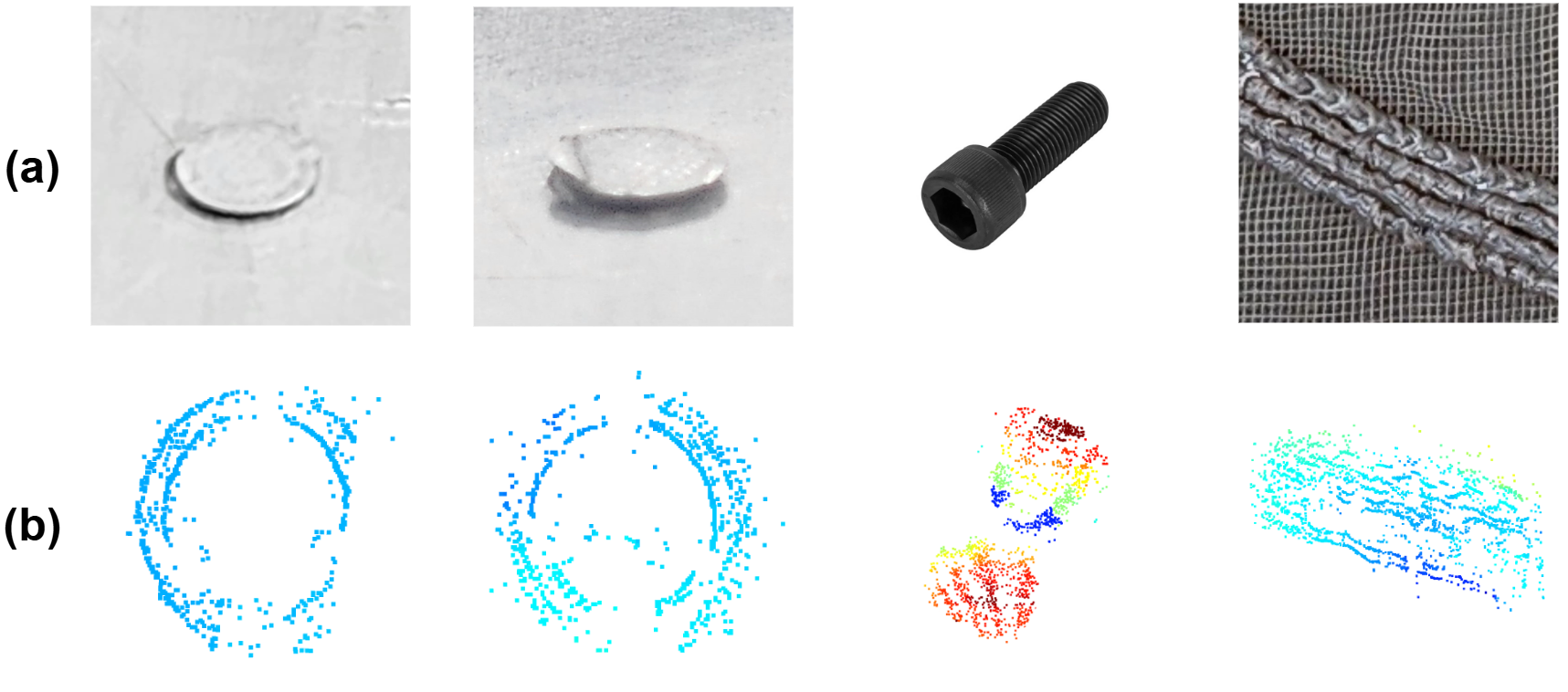}
\caption{\revisionTwo{Qualitative reconstruction results. \textbf{(a)} Physical objects: a flush rivet head, a rivet head with a defect, an M2 screw, and a textured fabric. \textbf{(b)} Corresponding 3D point cloud reconstructions. Color represents change in depth using JET colormap}}
\label{fig:qualitative_extra}
\end{figure}

To further assess the sensor's qualitative performance and generalization, we tested it on four additional objects with diverse materials and surface complexities, as shown in Figure \ref{fig:qualitative_extra}. The scanned objects include a healthy flush rivet, a defective rivet with a bent head, an M2 screw with fine threads, and a textured fabric. Comparing the healthy (Fig. \ref{fig:qualitative_extra} - Column 1) and defective (Fig. \ref{fig:qualitative_extra} - Column 2) rivets clearly demonstrates the sensor's defect detection capability: the healthy rivet yields a uniform depth map indicating flushness, while the defective rivet reveals a distinct height deviation corresponding to the bent head. The resulting point clouds in all cases exhibit strong geometric consistency with the physical objects. These results demonstrate the sensor's robustness in reconstructing a variety of complex geometries beyond the tested 3D printed shapes and potential for industrial surface inspection applications.

\subsection{Large Surface Reconstruction}
\label{subsec:large_surface_recon}
\begin{figure*}[t] \centering 
\includegraphics[width = \linewidth]{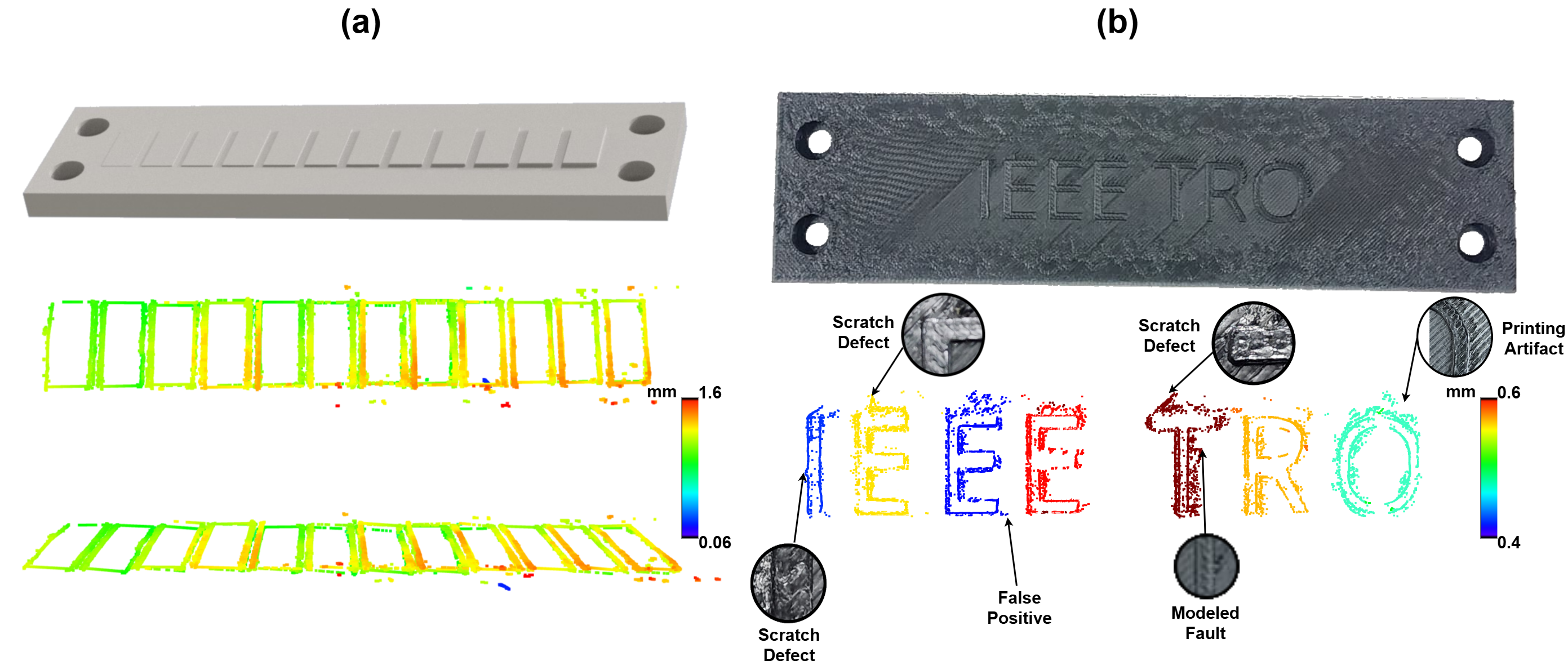}
\caption{\revised{Examples of large surface reconstruction. \textbf{(a)} 3D printed varying-height line array (top) and its corresponding reconstruction (bottom). \textbf{(b)} 3D printed letter plate with varying-height letters and induced defects (top) and its reconstruction (bottom).} \revisionTwo{Zoomed insets highlight manually induced defects either with scratches or intentionally modeled fault as well as examples of a false positive and printing artifacts.}} \label{fig:surface_recon} \end{figure*}

The large surface reconstruction capabilities of our roller were tested with two 3D printed plates, shown in Figure \ref{fig:surface_recon}. The final point cloud for each is generated by aligning results from multiple depth maps as done by Zhang et al. \cite{zhang_gelroller_2024}. The multiple depth maps are aligned using ICP; since the speed and hence the translation of the roller are known from the robotic controller, the translation along the rolling direction is used as the initial pose for ICP. \revisionTwo{It is important to note that due to the high repeatability of the UR10 robot and the short travel distances in our experiments, the initial pose derived from proprioception is already highly accurate. Quantitatively, the ICP step provided a maximum MAE reduction of approximately 0.0002 mm compared to using proprioception alone in our tests. While this refinement is minimal, the step is retained as a standard practice consistent with prior work \cite{zhang_gelroller_2024} to ensure the pipeline's robustness and generalizability for larger-area scans where robot drift could become a factor. The robot's motion error within a single 20 ms reconstruction window is negligible compared to other error sources and does not significantly impact the precision of an individual depth map.}

The test objects' reconstruction accuracy was averaged over five trials as was done for the single object reconstruction. The first test object was a line array with varying heights (0.1-1.2 mm at intervals of 0.1 mm), for which our sensor achieved a resulting MAE of 0.0638 $\pm$ 0.0055 mm across the whole surface (see Figure \hyperref[fig:surface_recon]{\ref*{fig:surface_recon}a}). The second object was a letter plate, designed to benchmark against prior work in letter reconstruction like GelRoller \cite{zhang_gelroller_2024} and to test defect detection  (see Figure \hyperref[fig:surface_recon]{\ref*{fig:surface_recon}b}). The letters on this plate have small, varying heights (ranging from 0.4 mm to 0.6 mm). \revisionTwo{Crucially, we focused our validation on specific induced defects: manually added scratches (seen on 'I' and the first 'E', and on top of the 'T') and a modeled fault added to the CAD design (on the side of the 'T'). As shown in the zoomed insets of Figure \hyperref[fig:surface_recon]{\ref*{fig:surface_recon}b}, our sensor successfully captured these induced defects. Additionally, the sensor detected unintentional printing artifacts, such as the diagonal layer lines on top of the second 'E' and on top of the 'O'. Notably, these artifacts were only reconstructed where their physical depth exceeded the sensor's sensitivity threshold; shallower printing lines in other regions did not deform the elastomer sufficiently to be detected, causing them to be naturally filtered out. Moreover, due to the noisy nature of event generation, some false positives are observed in the point cloud that could be misinterpreted as defects (e.g: right of second 'E'). Further discussion regarding false positive mitigation and detection of shallow tactile features is present in Section \ref{sec:discussion}}. For the letter plate, our sensor achieved an MAE of 0.0596 ± 0.0035 mm.

As aforementioned, speed is critical for large surface reconstruction using tactile sensing. Our roller exhibits, to the best of our knowledge, the highest measured speed at \textbf{0.5 m/s} or \textbf{500 mm/s} for tactile scanning with an MAE below \textbf{100 microns}. This is \textbf{11 times faster} than the maximum reported speed for rolling belt tactile sensors (45 mm/s) \cite{mirzaee_gelbelt_2025} and \textbf{45 times faster} than the highest speed for roller tactile sensors (11 mm/s) \cite{Cao2023}.

Table \ref{tab:sensor_comparison} summarizes the performance trade-offs for different continuous tactile sensing approaches based on the provided documents. Our roller significantly advances the state-of-the-art in terms of speed, achieving 500 mm/s while maintaining a competitive 3D reconstruction accuracy (\revised{62.0 $\mu m$ MAE}). GelBelt demonstrates the potential for very high accuracy ($\sim$13.5 $\mu m$ average MAE, measured against a high-resolution GelSight Max scan as ground truth for specific defects) but operates at a considerably lower maximum speed of 45 mm/s \cite{mirzaee_gelbelt_2025}. The GelRoller paper focused on a novel self-supervised reconstruction method, achieving 175 µm MAE in single-frame tests, but did not report a maximum scanning speed \cite{zhang_gelroller_2024}. TouchRoller operated at the slowest speed (11 mm/s) and was designed for 2D texture mapping, not 3D reconstruction \cite{Cao2023}.

\begin{table}[h]
\caption{Comparison between Tactile Belt/Roller Sensor Performance}
\label{tab:sensor_comparison}
\begin{center}
{

\resizebox{\linewidth}{!}{
\large
\begin{tabular}{|c|c|c|}
\hhline{===}
\multicolumn{1}{c}{Sensor} & \multicolumn{1}{c}{Max Reported Speed ($mm/s$)} & \multicolumn{1}{c}{MAE (\(\mu m\))} \\
\hhline{===}
GelBelt             & \underline{45}  & \textbf{\(\sim\)13.5} \\
\hline
GelRoller           & --- & 175 \\
\hline
TouchRoller         & 11  & --- \\
\hline
Ours & \textbf{500} & \underline{\revised{62.0}} \\
\hline
\end{tabular}
}
}
\end{center}
\end{table}

\subsection{High-Speed Tactile Braille Reading}
\begin{figure}[thpb]
    \centering
        \includegraphics[width=\linewidth]{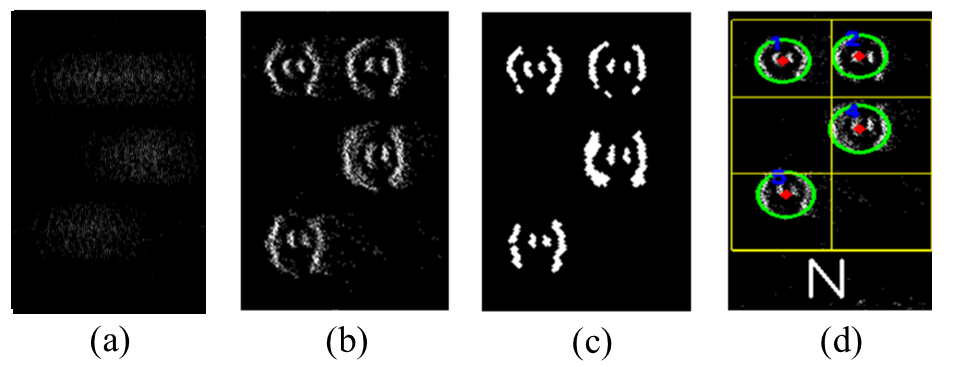}

    \caption{Braille feature recognition pipeline applied on single letter ROI. \textbf{(a)} Direct accumulation of events \textbf{(b)} Motion compensated image of letter. \revised{(c)} Denoised and processed letter image. \revised{(d)} Hough transform applied and corresponding letter identified.}
    \label{fig:braille_pipeline}
\end{figure}

The large surface tactile scanning of the roller was also tested with regards to feature detection and recognition. Two braille plates were scanned with standard braille sizing: semi sphere of radius of 0.635 mm and spacing of 2.54 mm. \revised{To obtain the Braille character recognition results, we processed the generated IWE, \(I_{\text{warp}}\), (from Section \ref{sec:braille}) with an application-specific pipeline. From \(I_{\text{warp}}\), we select a Region of Interest (ROI) corresponding to a single Braille cell location.} This ROI, illustrated in Figure \hyperref[fig:braille_pipeline]{\ref*{fig:braille_pipeline}b}, then undergoes an image processing pipeline to enhance dot visibility and reduce noise. The steps include: applying a bilateral filter for edge-preserving smoothing, using Contrast Limited Adaptive Histogram Equalization (CLAHE) to increase local contrast, performing binary thresholding to segment potential dots, and applying morphological opening and closing operations to eliminate salt-and-pepper noise.

Following pre-processing, the cleaned ROI (cf. Figure \hyperref[fig:braille_pipeline]{\ref*{fig:braille_pipeline}c}) is divided into the standard 2-column, 3-row Braille grid (6 squares). The Hough Circle Transform is then applied independently to each of the six squares to detect the presence or absence of a raised Braille dot. The detection results are encoded into a 6-bit binary vector, reading the grid from left-to-right, top-to-bottom, with '1' indicating a detected circle (dot) and '0' indicating its absence. This binary pattern is then compared against a Braille lookup table to identify the corresponding character (cf. Figure \hyperref[fig:braille_pipeline]{\ref*{fig:braille_pipeline}d}). This pipeline, illustrated in Figure \ref{fig:braille_pipeline}, enables the recognition of sequences of Braille characters directly from the high-speed tactile event data.

\begin{figure}[thpb]
    \centering
        \includegraphics[width=\linewidth]{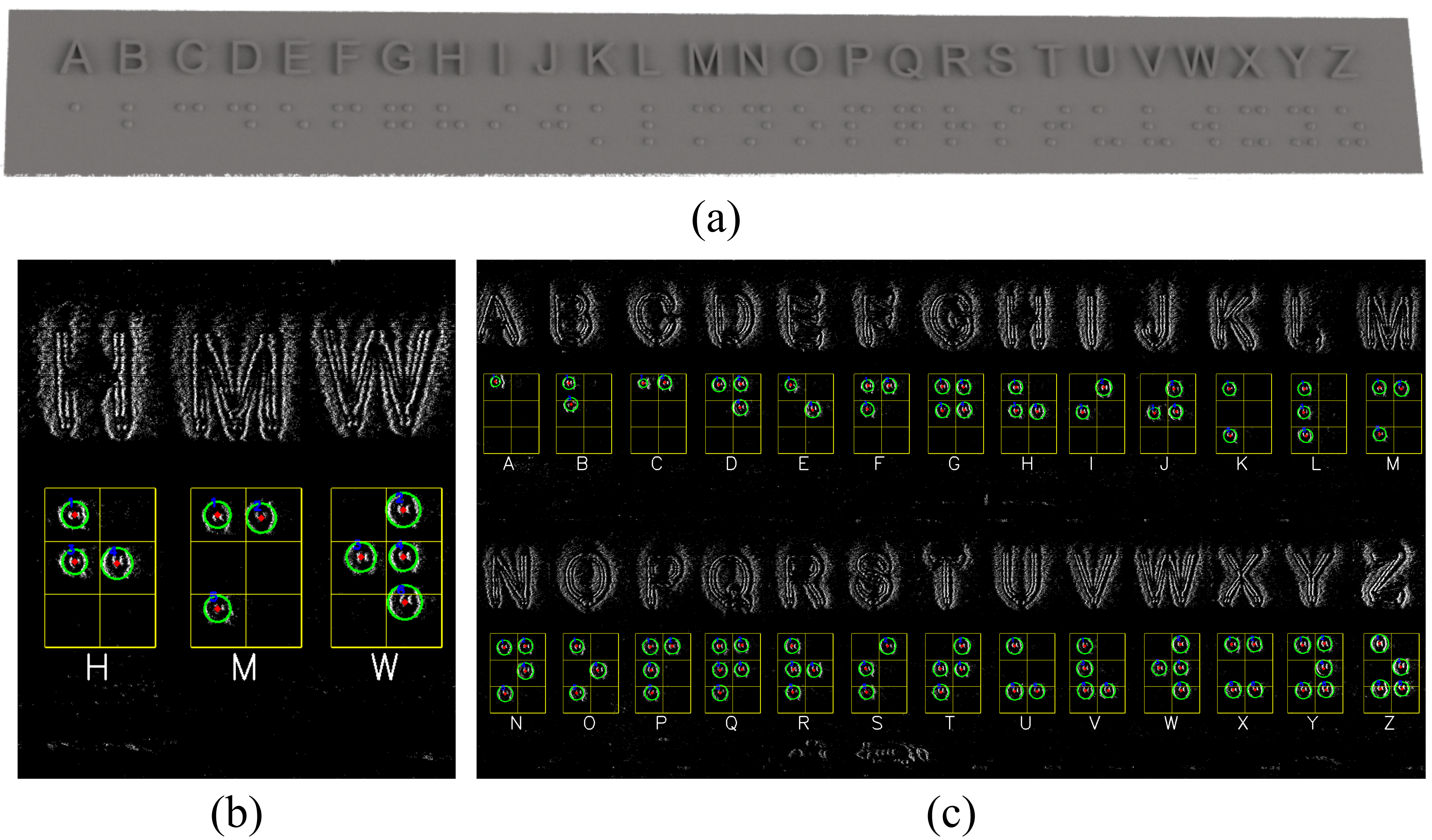} 
    \caption{High-speed braille reading. \revised{(a) Tactile braille plate (b) Close-up of a few detected braille letters (c) Full detected letters grid}}
    \label{fig:a_z_braille}
\end{figure}

The processing pipeline above was utilized to process the events into a frame and subsequently detect the letter in the frame. Each frame is then stitched to the next frame to form the complete words (as seen in Figure \ref{fig:graphical_abstract} and Figure \ref{fig:a_z_braille}). Due to the high temporal resolution (microsecond resolution) of the event camera, the recognition algorithm performs well at low and high speeds (0.05 to 0.5 m/s). \revised{For comparison, Potdar et al. \cite{potdar_high-speed_2024} demonstrated Braille reading at approximately 0.18 m/s with 87.5\% accuracy. \revisionTwo{Similarly, Xu et al. \cite{Xu_Braille_Neuromorphic} utilized Spiking Neural Networks (SNNs) to achieve 95.33\% accuracy in reliable Braille reading, though at a lower speed of 0.1 m/s.} Our system achieves the same task at 0.5 m/s, which is \textbf{2.6 times faster}, while maintaining full accuracy.} Furthermore, as Potdar et al. highlight, this tactile feature recognition pipeline can be extended for dynamic detection of surface textures \cite{potdar_high-speed_2024}.

\subsection{\revised{Processing Time and Computational Efficiency}}
\revised{
All experiments were run on a laptop CPU (13th Gen Intel Core i7-1365U) with offline processing of the recorded event data. The Braille reading pipeline, consisting of frame generation and Braille detection, achieved an average processing time of 19.26 ± 0.77 ms per frame (averaged over five runs). In comparison, the EMVS 3D reconstruction pipeline (from events to depth map) required 377.41 ± 7.96 ms per frame under the same conditions. This over 19× difference in processing time highlights the advantage of direct 2D feature recognition for tasks such as Braille reading, where the features of interest are small, well-defined, and do not require full 3D reconstruction for accurate detection. Although it's not explored in this work, the low processing time for the Braille pipeline indicates its strong potential for real-time operation. Conversely, the current CPU-based EMVS implementation is not real-time, as its processing time exceeds the 20 ms data acquisition window that was set for these experiments. However, leveraging a more powerful CPU and GPU acceleration for the highly parallelizable ray-casting and voting process could significantly improve performance and enable real-time 3D reconstruction.
}

\section{\revised{Discussion And Limitations}}
\label{sec:discussion}
Our experimental results demonstrate the significant capabilities of the proposed system. We achieved state-of-the-art continuous tactile scanning speeds of up to 0.5 m/s for 3D surface reconstruction, yielding Mean Absolute Errors (MAE) below 100 microns on large surface tasks. This speed represents a substantial leap, being approximately 11 times faster than reported belt-based sensors and 45 times faster than previous roller-based VBTSs. The BMA fusion technique proved crucial, reducing reconstruction MAE by an average of 25.2\% compared to standard EMVS. Furthermore, leveraging the sensor's high temporal resolution, we demonstrated Braille character recognition at 0.5 m/s, 2.6 times faster than the current benchmark for tactile Braille reading. These findings underscore the potential of neuromorphic vision to revolutionize high-speed tactile inspection, enabling applications previously infeasible due to the speed constraints of traditional methods, particularly in large-scale industrial quality control for aerospace or automotive components. \revised{This high-fidelity 3D reconstruction capability is crucial, as it allows for the direct comparison of a scanned surface against its reference CAD model to automatically detect and quantify volumetric defects.}

\revised{While the reliance on distinct edges and sharp features is a limitation of the underlying EMVS algorithm, it is important to contextualize this within the scope of industrial surface inspection. The primary goal in such applications is often the detection and quantification of high-frequency geometric anomalies, such as cracks, scratches, dents, weld seams, or the precise profiles of fasteners and drilled holes. These critical defects are inherently defined by the sharp edges and distinct features that our event-based system is specialized to capture with high fidelity. Therefore, while our method may not be ideal for reconstructing smooth, featureless surfaces, its strength aligns directly with the requirements for identifying the most common and critical types of defects in industrial quality control.}

\revised{An interesting observation from the calibration evaluation (Section \ref{subsec:calibration_evaluation}) is that the start reference view (\(t_s\)) consistently receives the highest weight, while the end view (\(t_e\)) receives the lowest. Intuitively, one might expect the midpoint view (\(t_m\)) to be most reliable, but several factors may contribute to this trend. First, the pose at \(t_s\) is derived directly from the robot state at the start of the window, whereas \(t_m\) and \(t_e\) rely on interpolation and may accumulate small kinematic or timing errors. Second, as the roller moves, tactile features first appear and generate high-contrast events near \(t_s\), potentially providing more stable features for EMVS before disocclusions occur. Finally, the DSI integrates events over the entire window, so errors later in the window may degrade the quality of the \(t_e\) viewpoint more than the \(t_s\) viewpoint. A more precise analysis of how the timing of the three reference windows affects reconstruction performance is also an interesting direction for future work.}

Despite the promising performance, the current approach exhibits limitations. The depth maps generated by the EMVS algorithm, relying on ray-counting from asynchronous and sparse event streams, are inherently semi-dense. While this works well for reconstructing objects with distinct edges or sufficient texture to generate events, it can struggle with accurately capturing smooth, continuous surface geometries where event generation might be sparse, potentially leading to incomplete or interpolated reconstructions. \revisionTwo{Additionally, edges aligned parallel to the rolling direction may generate fewer events due to minimal temporal contrast change \cite{neu_servoing}, resulting in minor reconstruction gaps at those specific orientations (e.g., the top/bottom edges of the rivet in Fig. 10). This effect is less apparent in large surface reconstructions, where slight kinematic variations over longer rolling distances reduce the likelihood of edges remaining perfectly parallel to the motion vector.} Moreover, the event camera used, the DVXplorer mini, has a resolution of 640x480 pixels (VGA). This inherently limits the spatial resolution that can be achieved by the tactile roller sensor. Currently, event camera technology is less mature than conventional camera technology. However, higher resolution sensors are emerging, such as the 1-megapixel event sensor presented by Finateu et al. \cite{Finateu2020}. Although these sensors are currently larger than the DVXplorer mini used here, integrating such improved hardware in future iterations would significantly enhance the spatial resolution of our sensor design.

\revisionTwo{Regarding the large surface reconstruction in Section \ref{subsec:large_surface_recon}, while the system successfully detected induced defects, distinguishing true defects from sensor noise (false positives) remains a challenge common to high-sensitivity tactile sensors. In Fig. \hyperref[fig:surface_recon]{\ref*{fig:surface_recon}b}, certain "defect-like" patterns are caused by signal noise. In future applications, these can be mitigated using spatial filtering based on the metric nature of our reconstruction; for instance, applying a size threshold to disregard unconnected depth clusters smaller than a specific physical dimension (e.g., $<0.2$ mm) relevant to the inspection tolerance. Furthermore, the detection threshold is physically coupled to the elastomer properties. In future applications, this sensitivity can be tuned by adjusting the skin hardness such as using softer skins to detect micro-features, or harder skins to filter out negligible surface roughness, allowing the sensor response to be matched to the desired defect definition.}

Future research will focus on addressing these limitations and expanding the sensor's capabilities. Firstly, we plan to investigate the integration of force and contact angle sensing and control. This could involve adding markers to the non-sensing sides of the roller, using their displacement within a learning-based framework to estimate applied force and roller orientation relative to the surface. Implementing closed-loop control based on this feedback would ensure consistent contact, which is crucial for maintaining high-quality tactile data acquisition, especially on non-planar surfaces. Secondly, to overcome the challenges associated with sparse depth maps from EMVS, we aim to explore alternative event-based reconstruction methods capable of generating denser surface information. Techniques such as event-based structured light measurement or event-based photometric stereo could potentially be adapted to the tactile roller geometry to directly estimate dense surface normals or depth maps from the event data. Achieving denser reconstructions would significantly enhance the sensor's ability to accurately measure complex, continuously varying surface geometries, broadening its applicability. \revised{Finally, a systematic investigation into the mechanical robustness and long-term durability of the elastomer is a crucial next step. Future work will involve quantifying the wear characteristics of the silicone skin over extended operational cycles and performing a mechanical analysis to determine the optimal elastomer parameters for roller-based applications}

\section{\revised{Conclusion}}

In this work, we introduced a novel high-speed neuromorphic vision-based roller tactile sensor designed for continuous large surface inspection, addressing the speed limitations inherent in conventional VBTS technologies. We presented the sensor's mechanical design, integrating an event camera within a rolling elastomer mechanism, and developed a 3D reconstruction pipeline. This pipeline adapts the event-based multi-view stereo (EMVS) method and incorporates a Bayesian Model Averaging (BMA) fusion strategy, combining depth information from multiple reference viewpoints within a temporal window to enhance accuracy and robustness. The proposed system demonstrates significant advances in tactile sensing speed and accuracy, achieving continuous scanning at 0.5 m/s with sub-100 micron precision, representing an order-of-magnitude improvement over existing continuous tactile sensing approaches. This breakthrough enables practical deployment for large-scale industrial surface inspection applications previously constrained by the speed limitations of conventional tactile sensors. \revised{By enabling rapid, high-fidelity 3D reconstruction of complex surfaces, this technology has the potential to significantly improve quality assurance workflows in industries such as aerospace, automotive, and precision manufacturing, reducing inspection times and increasing defect detection reliability.}

\section*{Acknowledgments}

We thank Mohammed Alshimmari for his help with the experiments. Moreover, thanks to Omer F. Mohamed for helping us come up with the manuscript title. This research was funded by Sandooq Al Watan under Project ID: KU-EXT-SWARD-2022-8434000486. The authors acknowledge the support of the Advanced Research and Innovation Center (KU-ARIC), a joint research center established by Khalifa University of Science and Technology and Aerospace Holding Company LLC, a wholly-owned subsidiary of Mubadala Investment Company PJSC.

\addtolength{\textheight}{-12cm}   

\bibliographystyle{IEEEtran}
\bibliography{prog}

\section*{Biography}
\vspace{-33pt}
\begin{IEEEbiographynophoto}{Akram Khairi} received the B.Sc. degree in Mechanical Engineering from New York University Abu Dhabi, Abu Dhabi, UAE. He is currently a Research Assistant at the Advanced Research \& Innovation Center (ARIC), Khalifa University, Abu Dhabi, UAE. His research interests include event vision-based tactile sensing, vision-based tactile sensing, event vision systems, and applications of AI in robotics.
\end{IEEEbiographynophoto}
\vspace{-33pt}
\begin{IEEEbiographynophoto}{Hussain Sajwani} received the B.Sc. degree in
applied mathematics and statistics from Khalifa
University, Abu Dhabi, United Arab Emirates, in
2021.
He is currently with the Advanced Research and
Innovation Center, Khalifa University. His research interests include event-based vision, neuromorphic computing, and tactile sensing.
\end{IEEEbiographynophoto}
\vspace{-33pt}
\begin{IEEEbiographynophoto}{Abdallah Mohammad Alkilany} received the B.Sc. degree in mechanical engineering from Al-Balqaa Applied University/Faculty of Engineering Technology, Amman, Jordan. He is currently a Research Assistant at the Advanced Research \& Innovation Center (ARIC), Khalifa University, Abu Dhabi, UAE. His research interests include additive manufacturing, mechanical design, robotics development, sensors, and modeling and simulation.
\end{IEEEbiographynophoto}
\vspace{-33pt}
\begin{IEEEbiographynophoto}{Laith AbuAssi} received the bachelor’s degree
in mechatronics engineering from Hashemite University, Jordan, in 2012. He was a Research
and Development Engineer and a Mechatronics
Systems Unit Supervisor with King Abdullah
Design and Development Bureau (KADDB). He is
currently a Senior Researcher with the Advanced
Research and Innovation Center (ARIC), United
Arab Emirates. He is a member with the Center for
Autonomous Robotic Systems, Khalifa University.
His research activities focus on the hardware and software design and
development of robotic systems that include space rovers, weapon stations,
medical robotics, industrial 4.0, the IoT, machine learning, and AI.
\end{IEEEbiographynophoto}
\vspace{-33pt}
\begin{IEEEbiographynophoto}{Mohamad Halwani} received the M.Sc. degree in
mechanical engineering from Khalifa University,
Abu Dhabi, United Arab Emirates, in 2021, where he is currently pursuing the Ph.D. degree in engineering with a focus on robotics.
His research interests include event-based vision, and vision-based tactile sensing and its applications in robotic manipulation.
\end{IEEEbiographynophoto}
\vspace{-33pt}
\begin{IEEEbiographynophoto}{Islam Mohamed Zaid} received the M.Sc. degree in aerospace engineering with Khalifa University, Abu Dhabi, United Arab Emirates, where he is currently pursuing the Ph.D. degree in engineering with a focus on robotics. His research interests include neuromorphic vision, tactile sensing, soft robotics, artificial intelligence, and numerical simulations.
\end{IEEEbiographynophoto}
\vspace{-33pt}
\begin{IEEEbiographynophoto}{Ahmed Awadalla} received the B.Sc. degree in Mechanical Engineering from Alexandria University, Alexandria, Egypt. He is currently a Research Associate with the Advanced Research and Innovation Center (ARIC), Khalifa University, Abu Dhabi, UAE. His research interests include advanced manufacturing, mechanical design, and vision-based tactile sensing.
\end{IEEEbiographynophoto}
\vspace{-33pt}
\begin{IEEEbiographynophoto}{Dewald Swart} received the B.Eng. degree in Mechanical Engineering and the M.Eng. degree in Electrical and Electronic Engineering from Stellenbosch University, South Africa. He is currently a Senior Robotics Engineer with Strata Manufacturing PJSC, Al Ain, UAE. He has over 12 years of experience in aerospace robotics and automation, having deployed systems for Boeing and Airbus assembly lines. His research interests include robot vision control, real-time machine vision, and high-precision robotic assembly.
\end{IEEEbiographynophoto}
\vspace{-33pt}
\begin{IEEEbiographynophoto}{Abdulla Ayyad} (Member, IEEE) received the M.Sc.
degree in electrical engineering from The University of Tokyo, Tokyo, Japan, in 2019.
He conducted research with the Spacecraft Control and Robotics Laboratory, The University of Tokyo. He is currently a Senior Research Associate with the Advanced Research and Innovation Center (ARIC), Khalifa University, Abu Dhabi, United Arab Emirates, working on several robot autonomy projects. His current research interests include the application of AI in the fields of perception, navigation, and control.
\end{IEEEbiographynophoto}
\vspace{-33pt}

\begin{IEEEbiographynophoto}{Yahya Zweiri} received the Ph.D. degree in
mechanical engineering from the King’s College
London, in 2003. He is currently a Professor with
the Department of Aerospace Engineering and the
Director of the Advanced Research and Innovation
Center, Khalifa University, United Arab Emirates.
Over the past two decades, he has actively participated in defense and security research projects at institutions, such as the Defense Science and Technology Laboratory, King’s College London, and the King Abdullah II Design and Development Bureau, Jordan. He has a prolific publication record, with over 130 refereed journals and conference papers as well as ten filed patents in USA and U.K. His primary research interests include robotic systems for challenging environments, with a specific emphasis on applied AI and neuromorphic vision systems.
\end{IEEEbiographynophoto}

\end{document}